\documentclass[twocolumn]{svjour3}  
\RequirePackage{fix-cm}

\smartqed  %
\usepackage{graphicx}
\usepackage{times}
\usepackage{epsfig}
\usepackage{epstopdf}
\usepackage{amsmath}
\usepackage{amssymb}
\usepackage{subfigure}
\usepackage{caption}
\usepackage{tabularx}
\usepackage[table]{xcolor}
\usepackage[normalem]{ulem}
\usepackage{algorithm}
\usepackage[noend]{algpseudocode}

\usepackage{mypackages}
\usepackage{rotating}
\usepackage{afterpage}

\newcommand{\AF}[1]{\textcolor{black}{#1}}

\def\datasetname{TREK-150} 
\def\datasetlink{\texttt{\url{https://machinelearning.uniud.it/datasets/trek150/}}}

\journalname{International Journal of Computer Vision}

\begin{document}

\title{Visual Object Tracking in First Person Vision}

\titlerunning{Visual Object Tracking in First Person Vision}

\author{Matteo Dunnhofer \and 
        Antonino Furnari \and 
        Giovanni Maria Farinella \and 
        Christian Micheloni}

\institute{
M. Dunnhofer, C. Micheloni \at 
Machine Learning and Perception Lab, University of Udine \\ 
Via delle Scienze 206, Udine, 33100, Italy \\
\email{\{matteo.dunnhofer, christian.micheloni\}@uniud.it} \and
A. Furnari, G. M. Farinella \at 
Image Processing Laboratory, University of Catania \\ 
Viale A. Doria 6, Catania, 95125, Italy \\
\email{\{furnari, gfarinella\}@dmi.unict.it} \and
*Pre-print, accepted 22 September 2022
}

\vspace{-5mm}

\maketitle

\abstract{
The understanding of human-object interactions is fundamental in First Person Vision (FPV). Visual tracking algorithms which follow the objects manipulated by the camera wearer can provide useful information to effectively model such interactions. In the last years, the computer vision community has significantly improved the performance of tracking algorithms for a large variety of target objects and scenarios. Despite a few previous attempts to exploit trackers in the FPV domain, a methodical analysis of the performance of state-of-the-art trackers is still missing. This research gap raises the question of whether current solutions can be used ``off-the-shelf'' or more domain-specific investigations should be carried out. This paper aims to provide answers to such questions. We present the first systematic investigation of single object tracking in FPV. Our study extensively analyses the performance of \rev{42} algorithms including generic object trackers and baseline FPV-specific trackers. The analysis is carried out by focusing on different aspects of the FPV setting, introducing new performance measures, and in relation to FPV-specific \rev{tasks}. The study is made possible through the introduction of \datasetname, a novel benchmark dataset composed of 150 densely annotated video sequences. Our results show that object tracking in FPV poses new challenges to current visual trackers. We highlight the factors causing such behavior and point out possible research directions. Despite their difficulties, we prove that trackers bring benefits to FPV downstream \rev{tasks} requiring short-term object tracking. We expect that generic object tracking will gain popularity in FPV as new and FPV-specific methodologies are investigated.

}

\section{Introduction}
First Person Vision (FPV) refers to the study and development of computer vision techniques considering images and videos acquired from a camera mounted on the head of a person -- which is referred to as the camera wearer. This setting allows machines to perceive the surrounding environment from a point of view that is the most similar to the one of human beings.
In the FPV domain, understanding the interactions between a camera wearer and the surrounding objects is a fundamental problem~\AF{\cite{EK55,Wang2020,liu2020forecasting,RULSTMpami,damen2016you,ragusa2020meccano,cai2016understanding,gberta_2017_RSS,bertasius2017unsupervised,Rhoi2020,Ego4D2021}}.
\AF{To model such interactions,}
the continuous knowledge of where \AF{an object of interest} is located inside the video frame \AF{is advantageous}.
Indeed, keeping track of object locations over time allows to understand which objects are moving, which of them are passively captured while not interacted, and how the user relates to the scene.

The benefits of tracking in FPV have been explored by a few previous works in the literature. %
\begin{figure*}[t]%
\centering
\includegraphics[width=.8\linewidth]{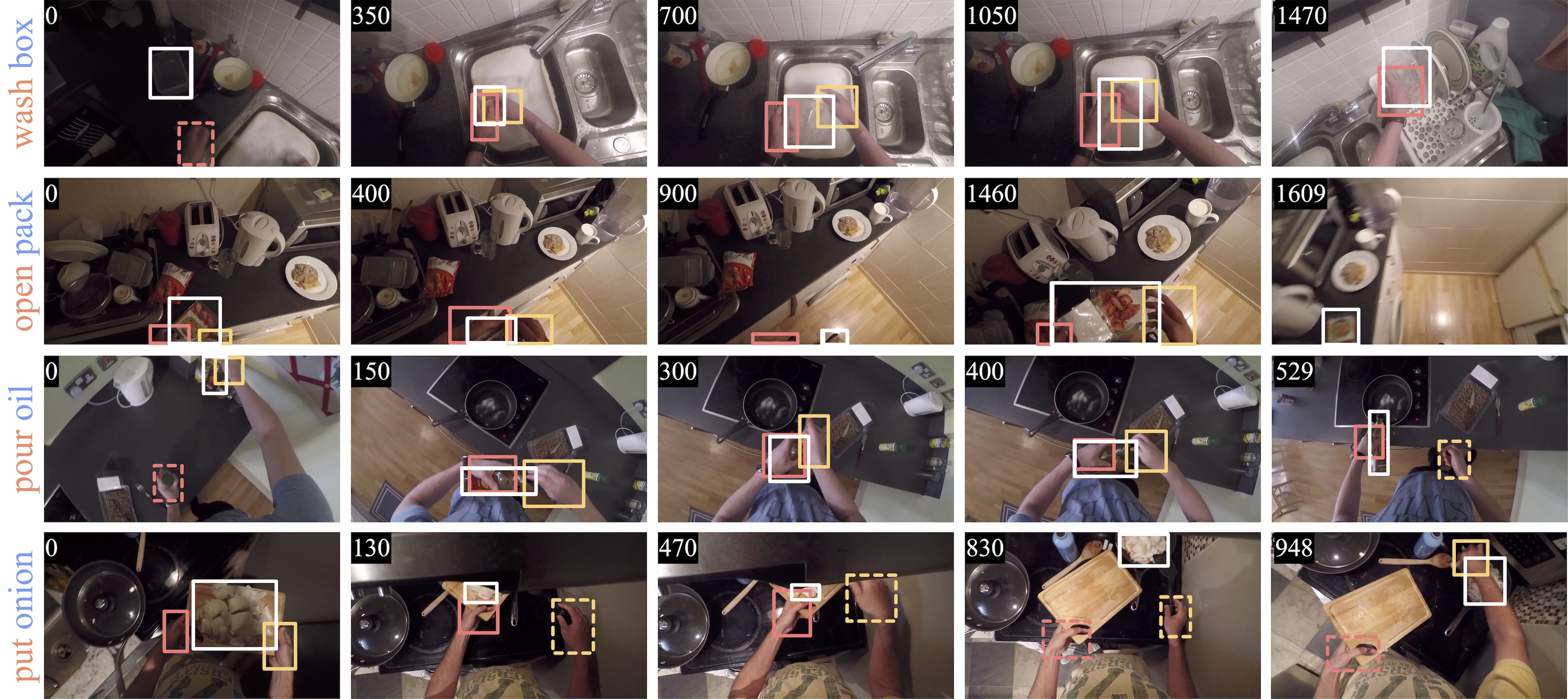}
\caption{In this paper, we study the problem of visual object tracking in the context of FPV. To achieve such a goal, we introduce a new benchmark dataset named \datasetname, of which some qualitative examples of sequences are represented in this Figure. In each frame, the white rectangle represents the ground-truth bounding box \AF{of} the target object. The orange and yellow boxes localize left and right hands respectively (plain lines indicate the interaction between the hand and the target). Each number in the top left corner reports the frame index. For each sequence, the action performed by the camera wearer is also reported (verb in orange, noun in blue). \AF{As can be noted, objects undergo significant appearance and state changes due to the manipulation by the camera wearer, which makes the proposed setting challenging for current trackers.}}
\label{fig:examples}
\end{figure*}
For example, visual trackers have been exploited in solutions 
to comprehend social interactions through faces~\cite{Aghaei2016,Aghaei2016icpr,Ego4D2021}, 
to improve the performance of hand detection for rehabilitation purposes~\cite{Visee2020}, to capture hand movements for action recognition~\cite{kapidis2019egocentric},
and to forecast human-object interactions through the analysis of hand trajectories~\cite{liu2020forecasting}.
Such applications have been made possible trough the development of customized tracking approaches to track specific target categories like people~\cite{Alletto2015,Nigam2017},
people faces~\cite{Aghaei2016,Ego4D2021}, or hands~\cite{kapidis2019egocentric,liu2020forecasting,Visee2020,Mueller2017,Han2020,Sun2010} from a first person perspective.

\AF{Despite the aforementioned attempts to leverage tracking in egocentric vision pipelines, the standard approach to \emph{generic-object continuous localisation} in FPV tasks still relies on}
detection models that evaluate video frames independently \cite{Wang2020,wu2019long,ma2016going,RULSTMpami,rodin2021predicting,sener2020temporal,EK55,EK100}.
This \AF{paradigm}
has the drawback of ignoring all the temporal information coming from the object appearance and motion contained in consecutive video frames. \AF{Also, it generally requires a higher computational cost due to the need to repeat the detection process in every frame.}
In contrast, visual object tracking aims to exploit past information about the target \AF{to}
infer its position and shape in the next frames of a video~\AF{\cite{ALOV,Maggio2011}}. 
This process can improve the efficiency of algorithmic pipelines because of the reduced computational resources needed, but most importantly because it allows to maintain the spatial and temporal reference to specific object instances. 

Visually tracking a \emph{generic object} in an automatic way introduces several different challenges  that include occlusions, pose or scale changes, appearance variations, and fast motion.
The computer vision community has 
\AF{made significant}
progress in the development of algorithms capable of tracking arbitrary objects in unconstrained scenarios affected by those issues. 
The advancements 
\AF{have been possible}
thanks to the development of new and effective \AF{tracking} principles \cite{MOSSE,KCF,SiamFC,ECO,DiMP,SiamGAT,Ocean,LTMU,Stark}, and to the careful design of benchmark datasets~\cite{OTB,UAV123,NfS,NUSPRO,LaSOT,GOT10k} and competitions \cite{VOT2017,VOT2019,VOT2020,VOT2021} 
that well represent the aforementioned challenging situations.
However, all these research endeavours have taken into account mainly the classic third person scenario in which the target objects are passively observed from an external point of view and where they do not interact with the camera wearer. 
It is a matter of fact that the nature of images and videos acquired from the first person viewpoint is inherently different from the type of image captured from video cameras set as on an external point of view.
As we will show in this paper, the particular characteristics of FPV, such as the interaction between the camera wearer and the objects as well as the proximity of the scene and the camera's point of view, cause the aforementioned challenges to occur with a \emph{different nature and distribution}, resulting in the persistent occlusion, significant scale and state changes of objects, as well as an increased presence of motion blur and fast motion (see Figure \ref{fig:examples}).

While the use cases of object tracking in egocentric vision are manifold and the benefit of tracking generic objects is clear as previously discussed, it is evident that visual object tracking is still not a dominant technology in FPV. Only very recent FPV pipelines are starting to employ generic object trackers \cite{TROI,Ego4D2021}, but a solution specifically designed to track generic objects in first person videos is still missing.
We think this lack of interest towards visual object tracking in FPV is mainly due to the \emph{limited amount of knowledge} present in the literature about the capabilities of current visual object trackers in FPV videos.
Indeed, this gap in the research 
opens many questions about the impact of the first person viewpoint on visual trackers: can the trackers available nowadays be used ``off-the-shelf''? How does FPV impact current methodologies? Which tracking approaches work better in FPV scenarios? What factors influence the most the tracking performance? What is the contribution of trackers in FPV? 
We believe that the particular setting offered by FPV deserves a dedicated analysis that is still missing in the literature, and we argue that further research on this problem cannot be pursued without a thorough study on the impact of FPV on tracking.

\AF{In this paper, we aim to}
extensively analyze the problem of visual object tracking in \AF{the} FPV \AF{domain} in order to answer the aforementioned questions. 
\AF{Given the lack of suitable benchmarks,} we follow the standard practice of the visual tracking community that suggests to build a curated dataset for evaluation \cite{OTB,TC128,UAV123,NUSPRO,NfS,VOT2019,CDTB}. Hence, we propose a novel visual tracking benchmark, \datasetname\linebreak(TRacking-Epic-Kitchens-150), which is obtained from the large and challenging FPV dataset EPIC-KITCHENS (EK) \cite{EK55,EK100}. \datasetname\ provides 150 video sequences which we densely annotated with the bounding boxes of a %
single target object the camera wearer interacts with. The dense localization of the person's hands and the interaction state between those and the target are also provided.
Additionally, each sequence 
\AF{has been}
labeled with attributes that identify the visual changes 
\AF{the object is undergoing, the class of the target object, }
as well as the action he/she is performing. 
By exploiting the dataset, 
we present an extensive and in-depth study of the accuracy and speed performance of 38 established generic object trackers and of \rev{4} newly introduced baseline FPV trackers.
We leverage standard evaluation protocols and metrics and propose new ones. This is done in order to evaluate the capabilities of the trackers in relation to specific FPV scenarios. Furthermore, we assess the trackers' performance by evaluating their impact on the FPV-specific downstream task of human-object interaction detection. 

In sum, the \emph{main contribution} of this manuscript is \textbf{the first systematic analysis of visual object tracking in FPV}. In addition to that, our study brings additional innovations: 
\begin{enumerate}[label=(\roman*)]
    \item the description and release of the new \datasetname\ dataset, which offers new challenges and complementary features with respect to existing visual tracking benchmarks;
    \item a new measure to assess the tracker's ability to maintain temporal reference to targets;
    \item a protocol to evaluate the performance of trackers with respect to a downstream task;
    \item \rev{four} FPV baseline trackers, \rev{two based on FPV object detectors and two combining such detectors with a state-of-the-art generic object tracker}.
\end{enumerate}

Our results show that FPV offers new and challenging tracking scenarios for the most recent \AF{and} accurate trackers \cite{LTMU,TrDiMP,ATOM,VITAL,ECO} and even for FPV trackers. 
We study the factors causing such performance and highlight possible future research directions. Despite the difficulties introduced by FPV, we prove that trackers bring benefits to FPV downstream \rev{tasks} requiring short-term object tracking such as hand-object interaction. 
Given our results and considering the potential impact in FPV, we expect that generic object tracking will gain popularity in this domain as new and FPV-specific methodologies are investigated.
\footnote{Annotations, trackers' results, and code are available at \datasetlink.}

\section{Related Work}

\subsection{Visual Tracking in FPV}
\label{sec:fpvtrackers}
There have been some attempts to tackle visual tracking in FPV.
Alletto et al.~\cite{Alletto2015} improved the TLD tracker~\cite{TLD} with a 3D odometry-based module to track people.
For a similar task, Nigam et al.~\cite{Nigam2017} proposed EgoTracker, a combination of the Struck~\cite{Struck} and MEEM~\cite{MEEM} trackers with a person re-identification module.
Face tracking was tackled by Aghaei et al. \cite{Aghaei2016} through %
\AF{a}
multi-object tracking approach 
\AF{termed}
extended-bag-of-tracklets.
Hand tracking was studied in 
\AF{several}
works\AF{~\cite{kapidis2019egocentric,Visee2020,Mueller2017,Han2020,Sun2010}}. Sun et al. \cite{Sun2010} developed a particle filter framework for hand pose tracking. Mueller et al. \cite{Mueller2017} instead proposed a solution based on an RGB camera and a depth sensor, while Kapidis et al. \cite{kapidis2019egocentric} and Vis\'ee et al. \cite{Visee2020} combined the YOLO \cite{YOLO} detector trained for hand detection with a visual tracker. The
\AF{former work}
used the multi-object tracker DeepSORT~\cite{DeepSORT}, whereas the 
\AF{latter}
employed the KCF \cite{KCF} single object tracker. Han et al. \cite{Han2020} exploited a detection-by-tracking approach on video frames acquired with 4 fisheye cameras.

All the aforementioned solutions focused on tracking specific targets (i.e., people, faces, or hands), and thus they \AF{are likely to} fail in generalizing to arbitrary target objects. 
Moreover, they have been validated on custom designed datasets, \AF{which limits the reproducibility of the works and the ability to compare them to other solutions}. 
\AF{In contrast,} we focus on the evaluation of algorithms \AF{for the generic object tracking task}.
\AF{We design our evaluation to be reproducible and extendable by releasing \datasetname, a set of 150 videos of different objects, which we believe will be useful to study object tracking in FPV.}
To the best of our knowledge, 
\AF{ours is the first attempt}
to evaluate systematically and in-depth generic object tracking in FPV.

\subsection{Visual Tracking for Generic Settings}
In recent years, there has been an increased interest in developing accurate and robust tracking algorithms for generic objects and domains.
Preliminary trackers were based on mean shift algorithms~\cite{Comanciu2000}, key-point~\cite{Matrioska}, part-based methods \cite{LGT,OGT}, or SVM~\cite{Struck} and incremental \cite{ross2008incremental} learning. Later, \AF{solutions based on} correlation filters gained popularity thanks to their processing speed~\cite{MOSSE,KCF,DSST,Staple,BACF}. More recently, \AF{algorithms based on} deep learning have been \AF{proposed} to extract efficient image and object features. This kind of representation has been used in deep regression networks~\cite{GOTURN,Dunnhofer2021ral}, online tracking-by-detection methods~\cite{MDNet,VITAL}, \AF{approaches} based on reinforcement learning~\cite{Yun2017,Dunnhofer2019}, deep discriminative correlation filters~\cite{ECO,ATOM,DiMP,PrDiMP,D3S,KYS}, \AF{trackers based on} siamese networks~\cite{SiamFC,SiamRPNpp,SiamMask,SiamGAT,Ocean,Dunnhofer2020MedIA}, and more recently in trackers built up on transformer architectures \cite{TransT,TrDiMP,Stark}. 
All these methods have been designed for tracking arbitrary target objects in unconstrained domains. However, no solution has been studied and validated on a number of diverse FPV sequences \AF{as} we propose in this paper.

\subsection{FPV Datasets and Tasks}
Different datasets are currently available in the FPV community for the study of particular tasks.
The CMU dataset \cite{delatorre2009} was introduced for studying the recognition of the actions performed by the camera wearer. Videos belonging to this dataset are annotated with labels expressing only the actions performed (up to 31) by the person, and they comprise around 200K frames. The EGTEA Gaze+ dataset \cite{li2018eye} extended the FPV scenarios represented in the previous dataset by providing 2.4M frames. Similarly as \cite{delatorre2009}, only labels for the actions performed by the camera wearer have been associated to the videos.
In addition to the action labels, the ADL dataset \cite{pirsiavash2012detecting} introduced around 137K annotations in the form of bounding boxes for the localization of the objects involved in the actions.  
Other than for the action recognition task, the MECCANO dataset \cite{ragusa2020meccano} was aimed to study active object detection and recognition as well as hand-object interaction. The dataset is designed to represent an industrial-like scenario and provides 299K frames, 64K bounding-boxes, 60 action labels, and 20 object categories.
The EPIC-KITCHENS dataset \cite{EK55,EK100} is currently one of the largest and most representative datasets available for vision-based tasks based on an egocentric point of view. It is composed of 20M frames and provides annotations for action recognition, action anticipation, and object detection.

Despite the extensive amount of labels for different FPV tasks, all the aforementioned  datasets \cite{pirsiavash2012detecting,ragusa2020meccano,EK55,EK100} do not offer annotations to study object tracking. This is because the available bounding boxes for the localization of objects are not relative to the specific instances of the objects but only to their categories. 
Such kind of annotations does not allow to distinguish different objects of the same category when these appear together in the images. 
Furthermore, such datasets provide only sparse annotations (typically at 1/2 FPS) and they do not provide tracking-specific annotations \cite{OTB,TrackingNet,VOT2017}. Hence, they cannot be used for an accurate and in-depth evaluation of trackers in FPV.
To the best of our knowledge, 
our proposed \datasetname\ dataset is the first tool that provides the chance of studying in-depth the visual object tracking task in the context of first-person viewpoint egocentric videos. 
In addition, with the release of dense annotations for the position of the camera wearer's hands, for the state of interaction between hands and the target object, and for the action performed by the camera wearer, 
\datasetname\ is suitable to analyze the visual tracking task in relation to all those FPV-specific tasks that require continuous and dense object localization (e.g. human-object interaction).

\begin{table}[h!]

\fontsize{8}{9}\selectfont
	\centering
	\caption{\rev{Statistics of the proposed \datasetname\ benchmark compared with other benchmarks designed for single visual object tracking evaluation. For the datasets marked with * we report the statistics of their test set. }}
	\label{tab:datasets}
	\setlength\tabcolsep{.16cm}
        \rowcolors{3}{tblrowcolor1}{tblrowcolor2}
	\rotatebox{90}{
	\begin{tabular}{l | c  c  c  c  c  c  c  c c c c | c }
		\toprule
		\multirow{2}{*}{Benchmark} & OTB-50 & OTB-100 & TC-128 & UAV123  & NUS-PRO & NfS  & VOT2019 & CDTB  & TOTB & GOT-10k* & LaSOT* & \multirow{2}{*}{\datasetname}\\
                    & \cite{OTB2013} & \cite{OTB} & \cite{TC128} & \cite{UAV123}  & \cite{NUSPRO} & \cite{NfS}  & \cite{VOT2019} & \cite{CDTB} & \cite{TOTB} &  \cite{GOT10k} & \cite{LaSOT} \\
		\midrule
		\# videos & 51 & 100 & 128 & 123 & 365 & 100 & 60 & 80 & 225 & 180 & 280 & 150 \\
		\# frames & 29K & 59K & 55K & 113K & 135K & 383K & 20K & 102K & 86K & 23K & 685k & 97K \\
		Min frames \AF{across videos} & 71 & 71 & 71 & 109 & 146 & 169 & 41 & 406 & 126 &  51 & 1000 & 161  \\
		Mean frames \AF{across videos} & 578 & 590 & 429 & 915 & 371 & 3830 & 332 & 1274 & 381 & 127 & 2448 & 649 \\
		Median frames \AF{across videos} & 392 & 393 & 365 & 882 & 300 & 2448 & 258 & 1179 & 389 & 100 & 2102 & 484 \\
		Max frames \AF{across videos} & 3872 & 3872 & 3872 & 3085 & 5040 & 20665 & 1500 & 2501 & 500 & 920 & 9999 & 4640 \\
		Frame rate & 30 FPS & 30 FPS & 30 FPS  & 30 FPS  & 30 FPS  & 240 FPS  & 30 FPS  & 30 FPS & 30 FPS & 10 FPS & 30 FPS & 60 FPS  \\
		\# target object classes & 10 & 16 & 27  & 9  & 8 & 17  & 30  & 23  & 15 & 84 & 70 & 34  \\
		\# sequence attributes & 11 & 11 & 11 & 12  & 12 & 9 & 6  & 13  & 12 & 6 & 14 & 17  \\
		Target absent labels & \xmark & \xmark & \xmark & \xmark & \xmark & \xmark & \cmark & \cmark & \cmark & \cmark & \cmark & \cmark \\
		Labels for the interaction with the target & \xmark & \xmark & \xmark & \xmark & \xmark & \xmark & \xmark & \xmark & \xmark &  \xmark & \xmark & \cmark \\
		FPV & \xmark & \xmark & \xmark & \xmark & \xmark & \xmark & \xmark & \xmark & \xmark & \xmark & \xmark & \cmark \\
		\# action verbs & n/a & n/a & n/a & n/a & n/a & n/a & n/a & n/a & n/a & n/a & n/a & 20 \\
		\bottomrule		
\end{tabular}
}

\end{table}

\subsection{Visual Tracking Benchmarks}
Disparate bounding-box level benchmarks are available today to evaluate the performance of single-object visual tracking algorithms. 
The Object Tracking Benchmarks (OTB) \linebreak OTB-50~\cite{OTB2013} and OTB-100 \cite{OTB} are two of the most popular benchmarks in the visual tracking community. They provide 51 and 100 sequences respectively, \AF{including} generic target objects like vehicles, people, faces, toys, characters, etc.
The Temple-Color 128~(TC-128) dataset \cite{TC128} comprises 128 videos that were acquired for the evaluation of color-enhanced trackers. 
The UAV123 dataset \cite{UAV123} was constructed to benchmark the tracking progress on videos captured by unmanned aerial vehicles (UAVs) cameras. The 123 videos included in this benchmark represent 9 different classes of target.
The NUS-PRO dataset~\cite{NUSPRO} contains 365 sequences and aims to benchmark human and rigid object tracking with targets belonging to one of 8 categories.
The Need for Speed (NfS) dataset~\cite{NfS} provides 100 sequences with a frame rate of 240 FPS. The aim of the authors was to benchmark the effects of frame rate variations on the tracking performance.
The VOT2019 benchmark~\cite{VOT2019} was the last iteration of the annual Visual Object Tracking challenge that required \linebreak bounding-boxes as target object representation. This dataset contains 60 highly challenging videos, with generic target objects belonging to 30 different categories. 
The Color and Depth Tracking Benchmark (CDTB) dataset \cite{CDTB} offers 80 RGB sequences paired with a depth channel. This benchmark aims to explore the use of depth information to improve tracking. 
The Transparent Object Tracking Benchmark (TOTB) \cite{TOTB} provides 225 videos of transparent target objects, and has been introduced to study the robustness of trackers to the particular appearance of such kind of objects.

Following the increased development of deep learning-based trackers, large-scale generic-domain tracking datasets have been recently released~\cite{TrackingNet,GOT10k,LaSOTijcv}. These include more than \AF{a} thousand videos normally split into training and test subsets. The evaluation protocol \AF{associated with} these sets requires the evaluation of the trackers after \AF{they have been trained on}
the provided training set.

Even though all the presented benchmarks offer various tracking scenarios, 
and some of them may include videos acquired from a first person point of view, 
no one was specifically designed for tracking in FPV. 
Moreover, since in this paper we aim to benchmark the performance of visual object trackers regardless of their approach, we follow the practice of \AF{previous works} \cite{OTB,TC128,UAV123,NUSPRO,NfS,VOT2019,CDTB,TOTB} and set up a well representative and described dataset for evaluation.
We believe that \datasetname\ is useful for the tracking community because it offers different tracking situations 
and new target object categories 
that are not present in other tracking benchmarks.

\section{The \datasetname\ Benchmark}
In this section, we describe \datasetname, the novel dataset proposed for the study of the visual object tracking task in FPV.
\footnote{Further motivations and details about \datasetname\ are given in the supplementary document.\label{footref}}
\datasetname\ is composed of 150 video sequences.
In each sequence, a single target object is labeled with a bounding box which encloses the appearance of the object in each frame in which the object is visible (as a whole or in part). 
Every sequence is additionally labeled with attributes describing the visual variability of the target and the scene in the sequence.
To study the performance of trackers in the setting of human-object interaction, we provide bounding box localization of hands and labels for their state of interaction with the target object. Moreover,
\AF{two additional verb and noun attributes are provided to indicate the action performed by the person and the class of the target, respectively.}
Some qualitative examples of the video sequences with the relative annotations are shown in Figure \ref{fig:examples}. Table \ref{tab:datasets} \AF{reports} key statistics of our dataset in comparison with \AF{existing} tracker evaluation benchmarks.
It is worth noticing that the proposed dataset is competitive in terms of size with respect to the evaluation benchmarks available in the visual (single) object tracking community.

We remark that \datasetname\ has been designed for the \emph{evaluation} of visual tracking algorithms in FPV regardless of their methodology. Indeed, in this paper, we do not aim to provide a large-scale dataset for the development of deep learning-based trackers. Instead, our goal is to assess the impact of the first-person viewpoint on current trackers.
To achieve this goal we follow the standard practice in the visual object tracking community \cite{OTB,TC128,UAV123,NUSPRO,NfS,VOT2019,CDTB,TOTB} that suggests to set up a small but \emph{well described dataset}
to benchmark the tracking progress.

\subsection{Data Collection}
\paragraph{Video Collection}
The videos contained in \datasetname\ have been sampled from EK \cite{EK55,EK100}, \AF{which is a public, large-scale, and diverse dataset of egocentric videos focused on human-object interactions in kitchens.}
This is currently one of the largest datasets for understanding human-object interactions in FPV.
Thanks to its dimension, EK provides a significant amount of diverse interaction situations between various people and several different types of objects. Hence, it allows us to select suitable disparate tracking sequences that reflect the common scenarios tackled in FPV tasks.
\AF{EK} offers videos annotated with the actions performed by the \AF{camera wearer} in the form of temporal bounds and verb-noun labels.
\AF{The subset of EK known as EK-55 \cite{EK55} also contains sparse bounding box references of manipulated objects annotated at 2 frames per second in a temporal window around each action.}
We exploited such a feature 
to obtain a suitable pool of video sequences interesting for object tracking. Particularly, we cross-referenced the original verb-noun temporal annotations of EK-55 to the sparse bounding box labels. This allowed to select sequences in which the camera wearer manipulates an object \rev{during an action}.
Each sequence is composed of the video frames contained \AF{within the temporal bounds of} the action, extracted \AF{at} the original 60 FPS frame rate and at the original \AF{full HD} frame size~\cite{EK55,EK100}. 
\AF{From the initial pool, we selected 150 video sequences which were characterized by attributes such as scale changes, partial/full occlusion and fast motion, which are commonly considered in standard tracking benchmarks~\cite{OTB,UAV123,TrackingNet,LaSOT,VOT2019}. 
The top part of Table~\ref{tab:attrdesc} reports the $13$ attributes considered for the selection.}

\begin{table}[t]
\fontsize{7}{7.5}\selectfont
\rowcolors{2}{tblrowcolor1}{tblrowcolor2}
	\centering
	\caption{Selected sequence attributes. The first block of rows describes attributes commonly used by the visual tracking community. The last four rows describe additional attributes \AF{introduced in this paper to characterize} FPV tracking sequences.}
	\label{tab:attrdesc}
	\begin{tabular}{m{3em} | m{27em} }
		\toprule
		Attribute & Meaning \\
		\midrule
		SC & \underline{Scale Change}: the ratio of the bounding-box area of the first and the current frame is outside the range [0.5, 2] \\
		ARC & \underline{Aspect Ratio Change}: the ratio of the bounding-box aspect ratio of the first and the current frame is outside the range [0.5, 2] \\
		IV & \underline{Illumination Variation}: the area of the target bounding-box is subject to light variation \\
		SOB & \underline{Similar Objects}: there are objects in the video of the same object category or with similar appearance to the target \\
		RIG & \underline{Rigid Object}: the target is a rigid object \\
		DEF & \underline{Deformable Object}: the target is a deformable object \\
		ROT & \underline{Rotation}: the target rotates in the video \\
		POC & \underline{Partial Occlusion}: the target is partially occluded in the video \\
		FOC & \underline{Full Occlusion}: the target is fully occluded in the video \\
		OUT & \underline{Out Of View}: the target completely leaves the video frame \\
		MB & \underline{Motion Blur}: the target region is blurred due to target or camera motion \\
		FM & \underline{Fast Motion}: the target bounding-box has a motion change larger than its size \\
		LR & \underline{Low Resolution}: the area of the target bounding-box is less than 1000 pixels in at least one frame \\
		\midrule
		HR & \underline{High Resolution}: the area of the target bounding-box is larger than 250000 pixels in at least one frame \\
		HM & \underline{Head Motion}: the person moves their head significantly thus causing camera motion \\
		1H & \underline{1 Hand Interaction}: the person interacts with the target object with one hand for consecutive video frames \\
		2H & \underline{2 Hands Interaction}: the person interacts with the target object with both hands for consecutive video frames \\
		\bottomrule		
\end{tabular}
\end{table}

\subsection{Data Labeling}
\label{sec:labeling}

\paragraph{Single Object Tracking}
In this study, we restricted our analysis to the tracking of a single target object per video. This has been done because in the FPV scenario a person interacts through his/her hands with one object at a time in general \cite{EK55,EK100}. If a person interacts with two objects at the same time those can be still tracked by two single object trackers. Moreover, focusing on a single object allows us to analyze better all the challenging and relevant factors that characterize the tracking problem in FPV.
We believe that future work could investigate the employment of multiple object tracking (MOT) \cite{MOT,Hota} solutions for a general understanding of the position and movement of all objects visible in the scene. We think the in-depth study presented in this paper will give useful insights for the development of such methods.

\begin{figure}[t]%
\centering
\includegraphics[width=.75\linewidth]{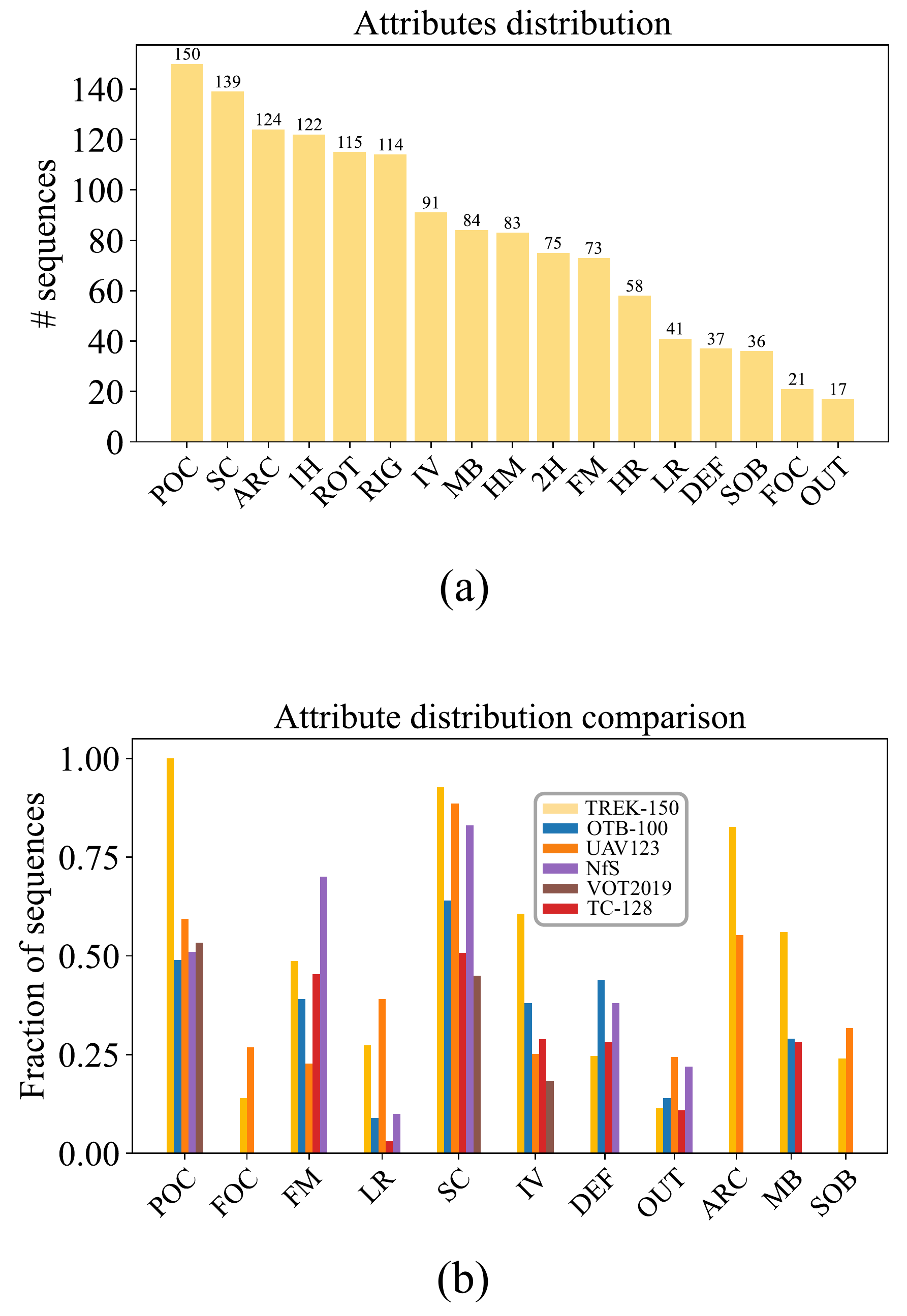}
\caption{(a) Distribution of the sequences within \datasetname\ with respect to the attributes used to categorize the visual variability happening on the target object and scene. (b) Comparison of \AF{the} distributions \AF{of} common sequence attributes across different benchmarks.}
\label{fig:attrdistributions}
\end{figure}

\begin{figure}[t]%
\centering
\includegraphics[width=.85\linewidth]{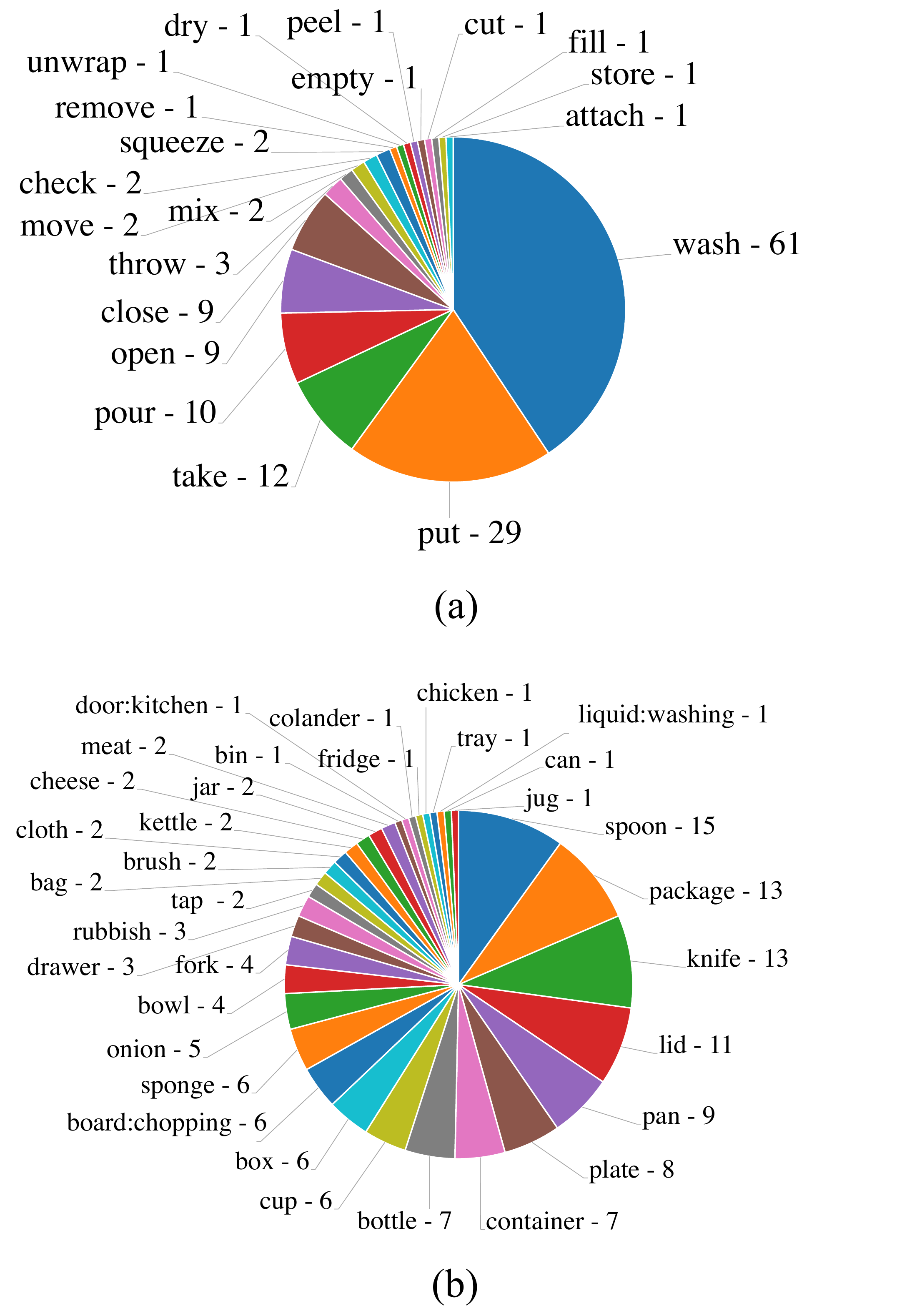}
\caption{Distributions of (a) action verb labels and (b) target object categories.}
\label{fig:vndistributions}
\end{figure}

\paragraph{Frame-level Annotations}
After selection, the 150 sequences \AF{were associated to only} %
3000 bounding boxes, due to the sparse \AF{nature of the} object annotations \AF{in} EK-55. 
Since it has been shown that visual tracking benchmarks require dense and accurate box annotations~\cite{VOT2019,UAV123,LaSOT,OxUvA}, we re-annotated the bounding boxes of the target objects on the 150 sequences selected. 
Batches of sequences were delivered to annotators (21 subjects) who were instructed to perform the labeling. 
Such initial annotations were then carefully checked and refined by a PhD student, and finally revised by an early-stage researcher and by two professors. This process produced 97296 frames labeled with bounding boxes related to the position and visual presence of objects the camera wearer is interacting with.
Following the initial annotations of EK-55, we employed axis-aligned bounding boxes to localize the target objects. 
This design choice is supported by the fact that such a representation is largely used in many FPV pipelines \cite{Furnari2017,RULSTMiccv,RULSTMpami,EK55,Kapidis2019,Visee2020,Shan2020}. Therefore, computing tracking metrics based on such representations allows us to correlate the results with those of object localization pipelines in FPV tasks, ultimately better highlighting the impact of trackers in such contexts.
Also, the usage of more sophisticated target representation would have restricted our analysis since the majority of state-of-the-art trackers output just axis-aligned bounding boxes \cite{MOSSE,DSST,KCF,MDNet,Staple,SiamFC,GOTURN,ECO,BACF,VITAL,STRCF,MCCTH,MetaTrackers,SiamRPNpp,SiamDW,ATOM,DiMP,SPLT,PrDiMP,GlobalTrack,SiamFCpp,LTMU,SiamBAN,Ocean,KYS,SiamGAT,TrDiMP,Stark,STMTrack},
and their recent progress on various benchmarks using such representation \cite{OTB,UAV123,NfS,CDTB,TOTB,TrackingNet,LaSOT,GOT10k} proves that it provides sufficient information for tracker initialization and consistent and reliable performance evaluation. Moreover, we point out that many of the objects commonly appearing in FPV scenarios are difficult to annotate consistently with more sophisticated target representations.%
We remark that the proposed bounding boxes have been carefully and tightly drawn around the visible parts of the objects.
Figure \ref{fig:annoquality} of the supplementary document shows some examples of the quality of the bounding-box annotations of \datasetname\ in contrast to the ones available in the popular OTB-100 tracking benchmark.

In addition to the bounding boxes for the object to be tracked, \datasetname\ provides per-frame annotations of the location of the left and right hand of the camera wearer and of the state of interaction happening between each hand and the target object. 
Interaction annotations consist of labels expressing which hand of the camera wearer is currently in contact with the target object (e.g., we used the labels LHI, RHI, BHI to express whether the person is interacting with the target by her/his left or right hand or with both hands). We considered an interaction happening even in the presence of an object acting as a medium between the hand and the target. E.g., we considered the camera wearer to interact with a dish even if a sponge is in between her/his hand and the dish. The fourth row of Figure \ref{fig:examples} shows a visual example of these situations.
These kinds of annotations have been obtained by the manual refinement (performed by the four aforementioned subjects) of the output given by the FPV hand-object interaction detector Hands-in-Contact (HiC) \cite{Shan2020}. 
\rev{In total, 166883 hand bounding boxes (82678 for the left hand, 84205 for the right hand) and 77993 interaction state labels (24466 for interaction with left hand, 16171 with right hand, 37356 with both hands) are present in TREK-150.}

\paragraph{Sequence-level Annotations}
The sequences have been also labeled considering 17 attributes which define the \AF{motion and visual appearance changes} the target object or the scene is subject to.
These are used to analyze the performance of the trackers under different aspects that may influence their execution.
\AF{The attributes employed in this study include 13 attributes used in standard tracking benchmarks \cite{OTB,TrackingNet,LaSOT}, plus 4 additional new ones (High Resolution, Head Motion, 1-Hand Interaction, 2-Hands Interaction) \linebreak which have been introduced in this paper to characterize sequences from FPV-specific point of views. The 17 attributes are \rev{defined} in Table~\ref{tab:attrdesc}.}
Figure~\ref{fig:attrdistributions}(a) reports the distributions of the sequences with respect to the 17 attributes, while
\AF{Figure~\ref{fig:attrdistributions}(b) compares the distributions of the most common attributes in the field in} \datasetname\ and in other well-known tracking benchmarks. 
Our dataset provides a larger number of sequences \AF{affected by partial occlusions}~(POC), \AF{changes in scale}~(SC) and/or aspect ratio~(ARC), motion blur~(MB), and illumination variation~(IV). 
These peculiarities are due to the particular first person viewpoint and to the human-object interactions which affect the camera motion and the appearance of objects. 
\AF{Based on the verb-noun labels of EK,} sequences were also associated to 
\AF{20 verb labels~(e.g., ``wash'' - see Figure~\ref{fig:examples}) and 34 noun labels indicating the category of the target object (e.g., ``box'').}
Figures~\ref{fig:vndistributions}(a-b) report the distributions of the videos with respect to 
verb and target object labels. As can be noted, our benchmark reflects the long-tail distribution of labels in EK \cite{EK55}.

\begin{table*}[t]
\fontsize{7.5}{8}\selectfont
	\centering
	\caption{\rev{Characteristics of the generic object trackers considered in our evaluation. We provide details about: the Image Representation employed by the trackers (Pixel column - \cmark if the tracker uses raw pixel intensity values; HOG column - \cmark if the tracker uses Histogram of Oriented Gradients; Color column - \cmark if the tracker uses Color Names or Intensity; CNN column - the Convolutional Neural Network backbone used);  the Matching Operation performed to find the target in sequence frames (CF column - \cmark if the tracker uses correlation filters; CC column - \cmark if the tracker uses the cross correlation; Concat column - \cmark if the tracker concatenates features; T-by-D column - \cmark if the tracker uses a tracking-by-detection approach;  Had column - \cmark if the tracker uses hadamard correlation; Tra column - \cmark if the tracker uses a transformer-based correlation). The \cmark\ symbol in the Model Update column expresses whether the tracker updates the target model during the tracking procedure. The next four columns report the category of tracking approach according to \cite{Lukezic2018me} ($\text{ST}_0$ column - short-term trackers without any re-detection mechanism; $\text{ST}_1$ column - short-term trackers without any re-detection mechanism but that estimate tracking confidence; $\text{LT}_0$ column - pseudo long-term trackers that do not detect failure and do not perform explicit re-detection; $\text{LT}_1$ column - long-term trackers that detect tracking failure and perform re-detection). The last column presents the datasets used to optimize the tracker in the offline training phase (I - ImageNet \cite{ImageNet}, IV - ILSVRC-VID \cite{ILSVRC15}, ID - ILSVRC-DET \cite{ILSVRC15}, C - COCO \cite{COCO}, Y - YouTube-BB \cite{Youtubebb}, YV - YouTube-VOS \cite{YTVOS}, A - ALOV \cite{ALOV}, O - OTB \cite{OTB}, V - VOT \cite{Kristan2016}, T - TrackingNet \cite{TrackingNet}, G - GOT-10k \cite{GOT10k}, L - LaSOT \cite{LaSOT}). }}
	\label{tab:trackers}
	\setlength\tabcolsep{.08cm}
	\rowcolors{3}{tblrowcolor1}{tblrowcolor2}
	\begin{tabular}{l | l  | c  c  c  c | c c c c c c | c | c c c c | c}
		\toprule
		\multirow{2}{*}{Tracker} & \multirow{2}{*}{Venue} & \multicolumn{4}{c|}{Image Representation}  & \multicolumn{6}{c|}{Matching Operation}  & Model & \multicolumn{4}{c|}{Class given by \cite{Lukezic2018me}} & Offline Training \\
		 & & Pixel & HOG & Color & CNN & CF & CC & Concat & T-by-D & Had & Tra & Update & $\text{ST}_0$  & $\text{ST}_1$ & $\text{LT}_0$ & $\text{LT}_1$ & Dataset \\
        
		\midrule
		
		MOSSE \cite{MOSSE} & CVPR 2010 & \cmark & & & & \cmark & & & & & & \cmark & \cmark & & & \\ 
		DSST \cite{DSST} & BMVC 2014 & \cmark & \cmark & & & \cmark & & & & & & \cmark & \cmark & & & & \\ 
		KCF \cite{KCF} & TPAMI 2015 & & \cmark & & & \cmark & & & & & & \cmark & \cmark & & & \\
		MDNet \cite{MDNet} & CVPR 2016 &  & & & VGG-M & & & & \cmark & &  & \cmark & & \cmark  & & & I, O, IV \\
		Staple \cite{Staple} & CVPR 2016 & & \cmark & \cmark & & \cmark & & & & &  & \cmark & \cmark &   & & \\
		SiamFC \cite{SiamFC} & ECCVW 2016 &  & & & AlexNet & & \cmark & & & &  &  & \cmark &  & & & G \\
		GOTURN \cite{GOTURN} & ECCV 2016 &  & & & AlexNet & & & \cmark & & &  &  & \cmark  &  & & & ID, A \\
		ECO \cite{ECO} & CVPR 2017 &  & & & VGG-M & \cmark &  & & & &  & \cmark  &  & \cmark  & & & \\
		BACF \cite{BACF} & ICCV 2017 & & \cmark & & & \cmark & & & & & & \cmark & \cmark & & & & \\
		DCFNet \cite{DCFNet} & ArXiv 2017 & & & & VGG-M & \cmark & & & & & & \cmark & \cmark & & & & \\
		VITAL \cite{VITAL} & CVPR 2018 & & & & VGG-M & & & & \cmark & &  & \cmark & & \cmark  & & & I, O, IV \\
		STRCF \cite{STRCF} & CVPR 2018 & & \cmark & & & \cmark & & & & & & \cmark & \cmark & & & & \\
		MCCTH \cite{MCCTH} & CVPR 2018 & & \cmark &  \cmark & & \cmark & & & & & & \cmark & \cmark & & & & \\
		DSLT \cite{DSLT} & ECCV 2018  &  & & & VGG-16 & & \cmark & & & &  & \cmark & \cmark &  & & & ID, IV, C \\
		MetaCrest \cite{MetaTrackers} & ECCV 2018 &  & & & VGG-M & \cmark &  & & & &  & \cmark  &  & \cmark  & & & I, ID, V\\
		SiamRPN++ \cite{SiamRPNpp} & CVPR 2019 &  & & & ResNet-50 & & \cmark & & & &  &  & \cmark &  & & & I, C, ID, IV, Y \\
		SiamMask \cite{SiamMask} & CVPR 2019 &  & & & ResNet-50 & & \cmark & & & &  &  & \cmark &  & & & I, ID, YV \\
		SiamDW \cite{SiamDW} & CVPR 2019 &  & & & ResNet-22 & & \cmark & & & &  &  & \cmark &  & & & I, ID, Y \\
		ATOM \cite{ATOM} & CVPR 2019  &  & & & ResNet-18 & \cmark &  & & & &  & \cmark  &  & \cmark  & & & I, C, L, T \\
		DiMP \cite{DiMP} & ICCV 2019 & & & & ResNet-50 & \cmark & & & & &  & \cmark  &  & \cmark  & & & I, C, L, T, G\\
		SPLT \cite{SPLT} & ICCV 2019 & & & & ResNet-50 & \cmark & & & & &  & \cmark  &  &  & & \cmark & IV, ID \\
		UpdateNet \cite{UpdateNet} & ICCV 2019 & & & & AlexNet & & \cmark  & & & &  & \cmark  & \cmark  &  & & & L \\
		SiamFC++ \cite{SiamFCpp} & AAAI 2020  &  & & & AlexNet & & \cmark & & & &  &  & \cmark &  & & & Y, ID, IV, C, G  \\
		GlobalTrack \cite{GlobalTrack} & AAAI 2020  &  & & & ResNet-50 & & & & & \cmark &  &  &  &  & \cmark & & C, G, L \\
		PrDiMP \cite{PrDiMP} & CVPR 2020 & & & & ResNet-50 & \cmark & & & & &  & \cmark  &  & \cmark  & & & I, C, L, T, G \\
		SiamBAN \cite{SiamBAN} & CVPR 2020 &  & & & ResNet-50 & & \cmark & & & &  &  & \cmark &  & & & IV, ID, C, G, L, Y\\
		D3S \cite{D3S} & CVPR 2020 &  & & & ResNet-50 & \cmark &  & & & &  &  & \cmark &  & & & YV \\
		LTMU \cite{LTMU} & CVPR 2020 &  & & & ResNet-50 & \cmark & \cmark & & \cmark & &  & \cmark & &  & & \cmark & I, L \\
		Ocean \cite{Ocean} & ECCV 2020 &  & & & ResNet-50 & & \cmark & & & &  &  & \cmark &  & & & Y, ID, IV, C, G  \\
		KYS \cite{KYS} & ECCV 2020 & & & & ResNet-50 & \cmark & & & & &  & \cmark  &  & \cmark  & & & T, L, G \\
		TRASFUST \cite{Dunnhofer2020accv} & ACCV 2020 & & & & ResNet-18 & & & \cmark  & & &  & & \cmark &   & & & G \\
		SiamGAT \cite{SiamGAT} & CVPR 2021 &  & & & ResNet-50 & & \cmark & & & &  &  & \cmark &  & & & Y, ID, IV, C, G \\
		TrDiMP \cite{TrDiMP} & CVPR 2021 & & & & ResNet-50 & \cmark & & & &  & \cmark & \cmark  &  & \cmark  & & & C, L, T, G \\
		LightTrack \cite{LightTrack} & CVPR 2021 &  & & & NAS & & \cmark & & & &  &  & \cmark &  & & & Y, ID, IV, C, G \\
		TransT \cite{TransT} & CVPR 2021 &  & & & ResNet-50 & &  & & & & \cmark & \cmark  &  &  & & & C, L, T, G \\
		STMTrack \cite{STMTrack} & CVPR 2021 &  & & & GoogLeNet & &  & \cmark & & &  & \cmark  & \cmark &  & & & G \\
		STARK \cite{Stark} & ICCV 2021 &  & & & ResNet-50 & &  &  & & & \cmark & \cmark  & & \cmark  & \cmark & & C, L, T, G \\
		KeepTrack \cite{KeepTrack} & ICCV 2021 & & & & ResNet-50 & \cmark & & & & &  & \cmark  &  & \cmark  & \cmark & & I, C, L, T, G \\
		\bottomrule		
\end{tabular}
\end{table*}

\section{Trackers}

\subsection{Generic Object Trackers}
Among the examined trackers, 
38 have been selected 
\AF{to represent}
different \AF{popular} approaches to generic-object visual tracking.
\AF{Specifically, in the analysis we} have included short-term trackers \cite{Lukezic2018me} 
based on both correlation-filters with hand-crafted features (MOSSE \cite{MOSSE}, DSST \cite{DSST}, KCF \cite{KCF}, Staple \cite{Staple}, BACF \cite{BACF}, DCFNet \cite{DCFNet}, STRCF \cite{STRCF}, MCCTH \cite{MCCTH}) 
\AF{and}
deep features (ECO \cite{ECO}, ATOM \cite{ATOM}, DiMP \cite{DiMP}, PrDiMP \cite{PrDiMP}, KYS \cite{KYS}, KeepTrack \cite{KeepTrack}). We also considered deep siamese networks (SiamFC \cite{SiamFC}, GOTURN \cite{GOTURN}, DSLT \cite{DSLT}, SiamRPN++ \cite{SiamRPNpp}, SiamDW \cite{SiamDW}, UpdateNet \cite{UpdateNet}, SiamFC++ \cite{SiamFCpp}, SiamBAN \cite{SiamBAN}, Ocean \cite{Ocean}, SiamGAT \cite{SiamGAT}, STMTrack \cite{STMTrack}), tracking-by-detection methods (MDNet \cite{MDNet}, VITAL \cite{VITAL}), as well as trackers based on target segmentation representations (SiamMask \cite{SiamMask}, D3S \cite{D3S}),  meta-learning~(MetaCrest \cite{MetaTrackers}),~fusion of trackers\linebreak(TRASFUST~\cite{Dunnhofer2020accv}), neural architecture search (LightTrack \cite{LightTrack}),  and transformers (TrDiMP \cite{TrDiMP}, TransT \cite{TransT}, \linebreak STARK \cite{Stark}).
The long-term \cite{Lukezic2018me} trackers SPLT \cite{SPLT}, GlobalTrack \cite{GlobalTrack}, and LTMU \cite{LTMU} have been also taken into account in the study. 
These kinds of trackers are designed to address longer target occlusion and out of view periods by exploiting an object re-detection module. 
All of the selected trackers 
\AF{are state-of-the-art approaches published}
between the years 2010-2021.
Table \ref{tab:trackers} reports detailed information about the 38 considered generic-object trackers regarding the: venue and year of publication; type of image representation used; type of target matching strategy; employment of target model updates; and category of tracker according to the classification of~\cite{Lukezic2018me}. 
For each tracker, we used the code publicly available and adopted default parameters 
\rev{in order to have a fair comparison between the different tracking methodologies (i.e., to avoid comparisons between trackers specifically optimized for TREK-150 and non-optimized trackers). The original hyper-parameter values lead to the best and most likely generalizable instances of all the trackers.}
The code was run on a machine with an Intel Xeon E5-2690 v4 @ 2.60GHz CPU, 320 GB of RAM, and an NVIDIA TITAN V GPU.

\begin{figure*}[t]%
\centering
\includegraphics[width=\linewidth]{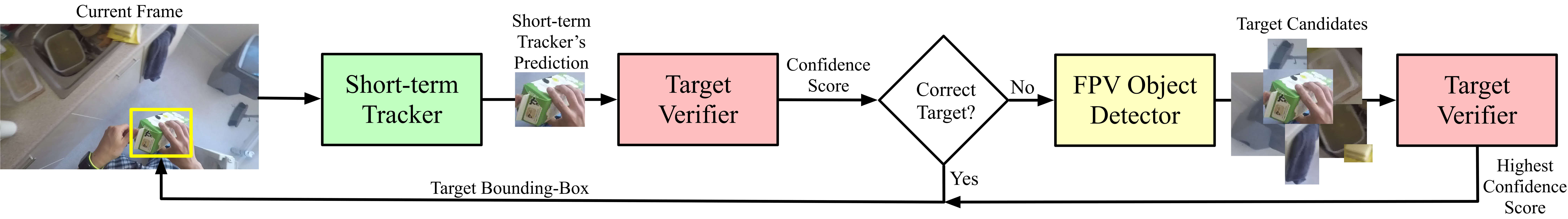}
\caption{Scheme of execution of the proposed FPV baseline trackers LTMU-F and LTMU-H based on LTMU \cite{LTMU}.}
\label{fig:fpvtracker}
\end{figure*}

\subsection{FPV Trackers}

Since there are no public implementations of the FPV trackers described in Section \ref{sec:fpvtrackers}, we introduce 4 new FPV-specific tracking baselines. 

\paragraph{\rev{TbyD-F/H}}
\label{sec:tbyd}
\rev{The first two FPV baselines build up on FPV-specific object detectors \cite{EK55,Shan2020}. Considering that they are popular approach for object localization in FPV and off-the-shelf FPV-trained instances are publicly available, we tested whether they can be used as na\"ive tracking baselines. To this end, we define a simple processing procedure which we found to work surprisingly well. At the first frame of a tracking sequence, the initial bounding box is memorized as current information about the target's object's position. Then, at every other frame, an FPV object detector is run to provide the boxes of all object instances present in the frame. As output for the current frame, the bounding-box having larger intersection-over-union (IoU) with the previously memorized box is given. If the detector does not output detections for a particular frame or none of its predicted boxes has IoU greater than 0, then the previously memorized box is given as output for the current frame. As object detectors, we used the EK-55 trained Faster-R-CNN \cite{FasterRCNN,EK55} and the Faster-R-CNN-based hand-object interaction detector HiC \cite{Shan2020}. The tracking baseline built upon the first detector is referred to as TbyD-F, while the one built on the second as TbyD-H.
} 

\paragraph{\rev{LTMU-F/H}}
We developed 2 other FPV-specific trackers in addition to the aforementioned ones. 
\rev{In this case, we wanted to combine the capabilites of generic object trackers with the FPV-specific object localization abilities of detectors \cite{EK55,Shan2020}.}
Particularly, the baselines combine the LTMU tracker \cite{LTMU} with FPV-specific object detectors. The first solution, referred to as LTMU-F, employs the Faster-R-CNN object detector trained on EK-55 \cite{EK55}, while the second, denoted as LTMU-H, uses the hand-object detector HiC~\cite{Shan2020}. 
These two trackers exploit the respective detectors as re-detection modules according to the LTMU scheme \cite{LTMU}.
For a better understanding, we briefly recap the processing procedure of the LTMU tracker~\cite{LTMU}. After being initialized with the target in the first frame of a sequence, at every other frame LTMU first executes a short-term tracker that tracks the target in a local area of the frame based on the target's last position. The patch extracted from the box prediction of the tracker is evaluated by an online-learned verification module based on MDNet \cite{MDNet}, which outputs a probability estimate of the target being contained in the patch. Such an estimate \rev{in companion with the tracker's predicted traget presence} are used to decide if the short-term tracker is tracking the target or not. If it is, its predicted box is given as output for the current frame. In the other case, a re-detection module is executed to look for the target in the whole frame. The re-detector returns some candidate locations which may contain the target and each of these is checked by the verification module. The candidate patch with the highest confidence is given as output and used as a new target location to re-initialize the short-term tracker. 
\rev{The verifier's output as well as the tracker's confidence are used to decide when to update the parameters of the first.}
\rev{Based on experiments, we used STARK \cite{Stark} as short-term tracker and the aforementioned FPV-based detectors as re-detection modules.}
For LTMU-F, such a module has been set to retain the first 10 among the many detections given as output, considering a ranking based on the scores attributed by the detector to each detection. If no detection is given for a frame, the last available position of the target is considered as a candidate location. For LTMU-H, we used the object localizations of the hand-object interaction detections given by the FPV version of HiC~\cite{Shan2020} as target candidate locations. HiC is implemented as an improved Faster R-CNN which is set to provide, at the same time, the localization of hands and interacted objects, as well as their state of interaction. As for LTMU-F, if no detection is given for a frame, the last available position of the target is considered as a candidate location.
For both detection methods, the original pre-trained models provided by the authors 
have been used.
The described setups, the common scheme of which is presented in Figure \ref{fig:fpvtracker}, give birth to two new FPV trackers that implement conceptually different strategies for FPV-based object localization and tracking. Indeed, the first solution aims to just look for objects in the scene, while the second one reasons in terms of the interaction happening between the camera wearer and the objects. 

The choice of using LTMU \cite{LTMU} as a baseline methodology stems from its highly modular scheme which makes it the most easily configurable tracker with state-of-the-art performance available today.
We took advantage of the commodity of a such framework to insert the FPV-specific modules described before.

\section{Evaluation Settings}
\label{sec:eval}

\subsection{Evaluation Protocols}
The protocols used to execute the trackers are described in the following.

\paragraph{One-Pass Evaluation} 
We employed the one-pass evaluation (OPE) protocol detailed in \cite{OTB} which implements the most realistic way to run a tracker in practice.
The protocol consists of two main stages: (i) initializing a tracker with the ground-truth bounding box \AF{of} the target in the first frame; (ii) letting the tracker run \AF{on}
 every \AF{subsequent} frame until the end of the sequence \AF{and record} predictions to be considered for the evaluation. To obtain performance scores for each sequence, predictions and ground-truth bounding boxes are compared according to some distance measure only in frames where ground-truths are present (ground-truth bounding boxes are not given for frames in which the target is fully occluded or out of the field of view). The overall scores
 are obtained by averaging the scores achieved for every sequence.
 
 \rev{The tracker initialization with the ground-truth is performed to evaluate the trackers in the best possible conditions, i.e. when accurate information about the target is given. In practical applications, such a user-defined information is generally unavailable. We expect this scenario to occur especially in FPV applications where object localization is obtained via detectors \cite{EK55,Shan2020}. Detectors predict bounding boxes with spatial noise (in the position and/or in the scale), and the initialization of trackers with such a noisy information could influence the tracking performance.
 Hence, to understand the impact of the initial box given by an object detector, we consider a version of the OPE protocol, referred to as OPE-D, where each tracker is initialized in the first frame in which the detector's prediction  has IoU $\geq$ 0.5 with the ground-truth box. From such a frame (that could be delayed in time with respect to the beginning of the sequence), each tracker is also run with the ground-truth box. The change in the metric values obtained after running the two modalities are used to quantify the impact of the initialization box. }
 
\begin{figure*}[t]%
\centering
\includegraphics[width=.9\linewidth]{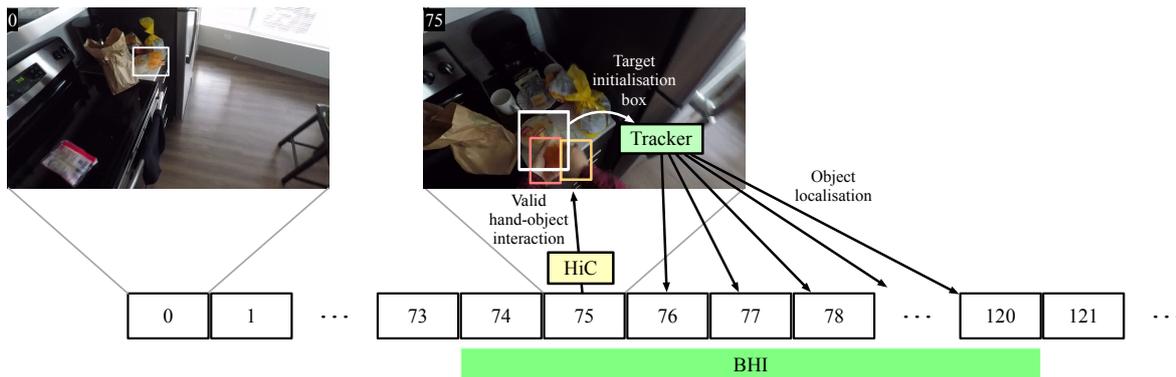}
\caption{Schematic visualization of the protocol designed to execute trackers in the context of a hand-object interaction (HOI) detection task. The HOI labels provided for \datasetname\ are used to consider sub-sequences of frames in which the camera wearer is interacting with the target object. In this picture, the labels BHI are employed to indicate that an interaction by both hands is happening in the frame range [74, 120]. 
On such sub-sequences, a systematic pipeline for HOI detection and tracking is run. The HOI detector HiC \cite{Shan2020} is first executed in every frame to obtain a valid HOI (in this example the first valid detection is obtained at frame 75). Once such an event is determined, the tracker is initialized with the bounding box given by HiC for the object involved in the interaction. The tracker is then run on all the subsequent frames to provide the reference to such an object. }
\label{fig:hicprotocol}
\end{figure*}

\paragraph{Multi-Start Evaluation} 
To obtain a more robust \linebreak evaluation~\cite{Kristan2016}, especially for the analysis over sequence attributes and action verbs, we employed the recent protocol proposed in~\cite{VOT2020}, which defines different points of initialization along a video. 
In more detail, for each sequence, different initialization points --called anchors-- separated by 2 seconds 
are defined. Anchors are always set in the first and last frames of a sequence. Some of the inner anchors are shifted forward by a few frames in order to avoid frames in which the target is not visible. A tracker is run on each of the sub-sequences yielded by the anchor 
either forward or backward in time depending on the longest sub-sequence the anchor generates. The tracker is initialized with the ground-truth annotation in the first frame of the sub-sequence and let run until its end. Then, as for the OPE, predicted and ground-truth boxes are compared to obtain performance scores for each sub-sequence. Scores for a single sequence are computed by averaging the scores of each sub-sequence weighted by their length in number of frames. Similarly, the overall scores for the whole dataset 
are obtained by averaging each sequence's score weighted by its number of frames.
 We refer to this protocol as multi-start evaluation (MSE). It allows a tracker to better cover all the situations happening in the sequences, ultimately leading to more robust evaluation scores.
 
\paragraph{\rev{Hand-Object Interaction Evaluation}}
\label{sec:evalfpv}
We also evaluated trackers in \rev{relation to} a video-based hand-object interaction (HOI) detection solution. \rev{This is done} in order to assess their direct impact on a downstream FPV-specific task. 
The aim of this problem is to determine when and where in the frames the camera wearer is interacting (e.g., by touching/manipulating) with an object with his/her hands.
\rev{Considering the requirement of generic object localization \cite{Shan2020}, we think a video-based configuration of such a problem to be a suitable task to exploit visual object trackers.}
To achieve the goal, we built a solution composed of a HiC instance \cite{Shan2020} to detect the hands and their state of interaction with an object and a visual tracker to maintain the reference to it. %
\rev{The HiC detector is run at every frame until it finds a valid HOI detection. Such an event is said to occur when the bounding box predictions for the hands have an IoU $\geq 0.5$ with the hand ground-truth boxes, the predicted interaction state is ``in contact'', and the object bounding box has an IoU $\geq 0.5$ with the ground-truth box \cite{Shan2020}. Then, the predicted object-related box is used to initialize the tracker, and for the subsequent frames, it is run to provide the localization of that object (that is the one involved in the interaction). A graphical representation of the execution of the described pipeline is given in Figure \ref{fig:hicprotocol}.}
Taking inspiration from the metric used by \cite{Shan2020} to evaluate HiC on static images, we quantify the performance of the proposed pipeline by the normalized count of frames in which the given HOI detection matches the ground-truth annotation available. Such matching is said to happen when the bounding box predictions for the hands have an IoU $\geq 0.5$ with the hand ground-truth boxes, the predicted interaction state is ``in contact'', and the object bounding box has an IoU $\geq 0.5$ with the ground-truth box \cite{Shan2020}.
For our experiments, we restricted the analysis of the solution on the sub-sequences contained in \datasetname\ in which an HOI is present.
These are determined by considering the sub-sequences of consecutive frames having the same interaction label (i.e., LHI, RHI, BHI). 
To obtain an overall performance score, which we refer to as Recall, we average the sub-sequence scores after having them weighted by the sub-sequence lengths in number of frames, in a similar fashion as we did to compute score in the MSE.
To evaluate the impact of visual trackers on this task, we switch the pipeline's tracker with each of the ones studied in this work. 
This experimental procedure gives us an estimate of the accuracy of the HOI detection system under configurations with different trackers. More interestingly, the proposed evaluation protocol allows also to build a ranking of the trackers based on the results of a downstream application. To the best of our knowledge, this setup brings a new way to assess the performance of visual object trackers.

\paragraph{Real-Time Evaluation} 
Since many FPV tasks such as object interaction \cite{damen2016you} and early action recognition \cite{RULSTMiccv}, or action anticipation \cite{EK55}, require real-time computation, we evaluate trackers in such a setting by following the instructions given in \cite{VOT2017,Li2020}. 
\rev{Explanations and results are given in the supplementary document.}

\subsection{Performance Measures}
\begin{figure}[!ht]%
\centering
\includegraphics[width=.975\columnwidth]{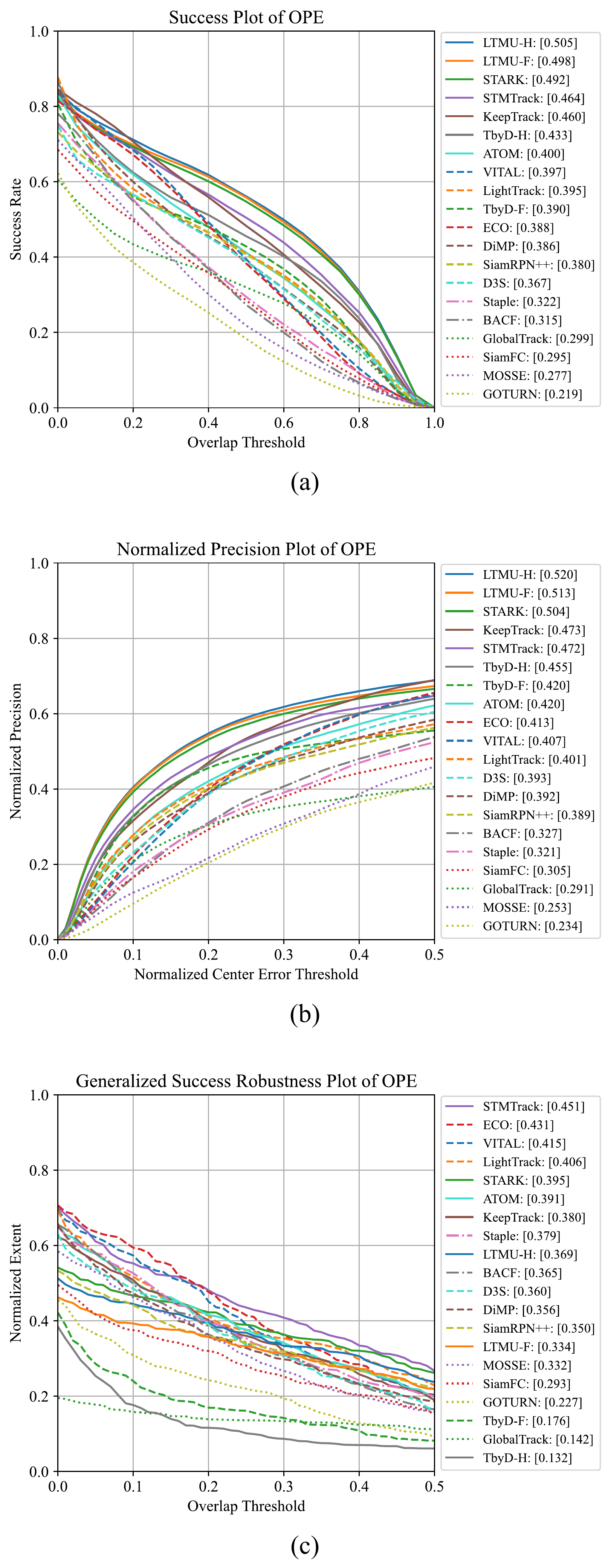}
\caption{Performance of 20 of the 42 selected trackers on the proposed \datasetname\ benchmark under the OPE protocol. \AF{In brackets, next to the trackers' names, we report the SS, NPS, and GSR values.}}
\label{fig:results}
\end{figure}

To quantify the performance of the trackers, we used different measures that compare trackers' predicted bounding boxes with the temporally aligned ground-truth boxes. 
To evaluate the overall \AF{localization accuracy of the trackers, we employ}
the success plot \cite{OTB}, which shows the percentage of \AF{predicted} boxes whose IoU with the ground-truth is larger than a threshold \AF{varied from 0 to 1} (Figure~\ref{fig:results}~(a)). 
We also use the normalized precision plot \cite{TrackingNet}, that \AF{reports}, \AF{for a variety of thresholds},
the percentage of boxes whose center points are within a given normalized distance from the ground-truth (Figure \ref{fig:results} (b)). 
\AF{As summary measures, we report the success score (SS) \cite{OTB} and normalized precision scores (NPS) \cite{TrackingNet}, which are computed as the Area Under the Curve (AUC) of the success plot and normalized precision plot respectively.}

Along with these standard metrics, we employ a novel plot which we refer to as generalized success robustness plot (Figure \ref{fig:results} (c)). For this, we take inspiration from the robustness metric proposed in \cite{VOT2020} which measures the normalized extent of a tracking sequence before a failure. 
We believe this aspect to be especially important in FPV as a superior ability of a tracker to maintain longer references to targets can lead to the better modeling of actions and interactions.
The original metric proposed in \cite{VOT2020} uses a fixed threshold of 0.1 on the bounding box overlap to detect a collapse of the tracker. Such a value was determined mainly to reduce the chance of cheating in the VOT2020 competition and it is not necessarily the case that such a value could work well for different tracking applications. 
To generalize the metric, we take inspiration from the success and normalized precision plots and propose to use different box overlap thresholds ranging in [0, 0.5] to determine the collapse. 
We consider 0.5 as the maximum threshold as higher overlaps are usually associated to positive predictions in many computer vision tasks. 
Overall, our proposed plot allows to assess the length of tracking sequences in a more general way that is better aligned with the requirements of different application scenarios including FPV ones.
\AF{Similarly to \cite{OTB,TrackingNet}, we use the AUC of the generalized success robustness plot to obtain an aggregate score which we refer to as generalized success robustness (GSR).}

\begin{figure}[t]%
\centering
\includegraphics[width=\columnwidth]{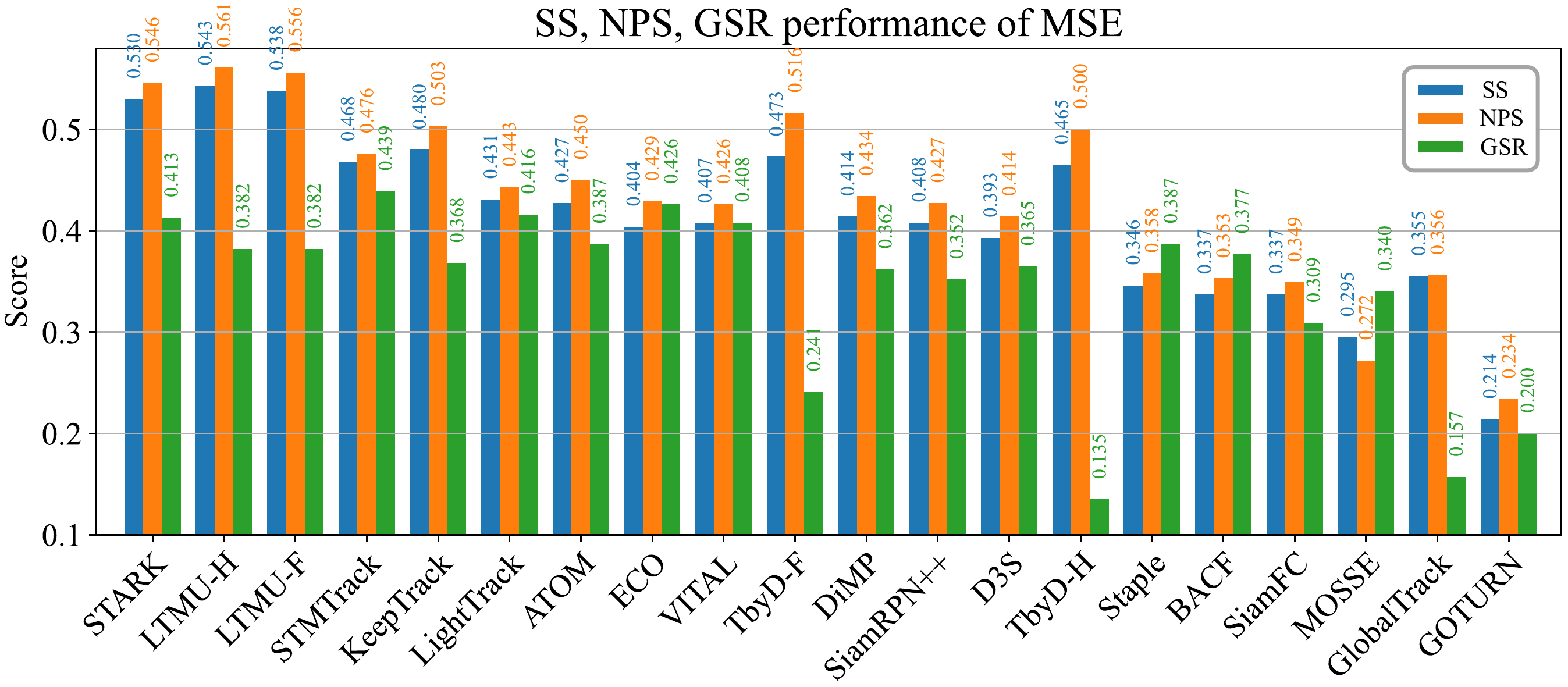}
\caption{SS, NPS, and GSR performance of 20 of the 42 benchmarked generic object trackers on the proposed \datasetname\ benchmark achieved under the MSE protocol. The trackers are ordered by the average value of their SS, NPS, GSR scores.}
\label{fig:resultsmse}
\end{figure}

\begin{figure*}[t]%
\centering
\includegraphics[width=\linewidth]{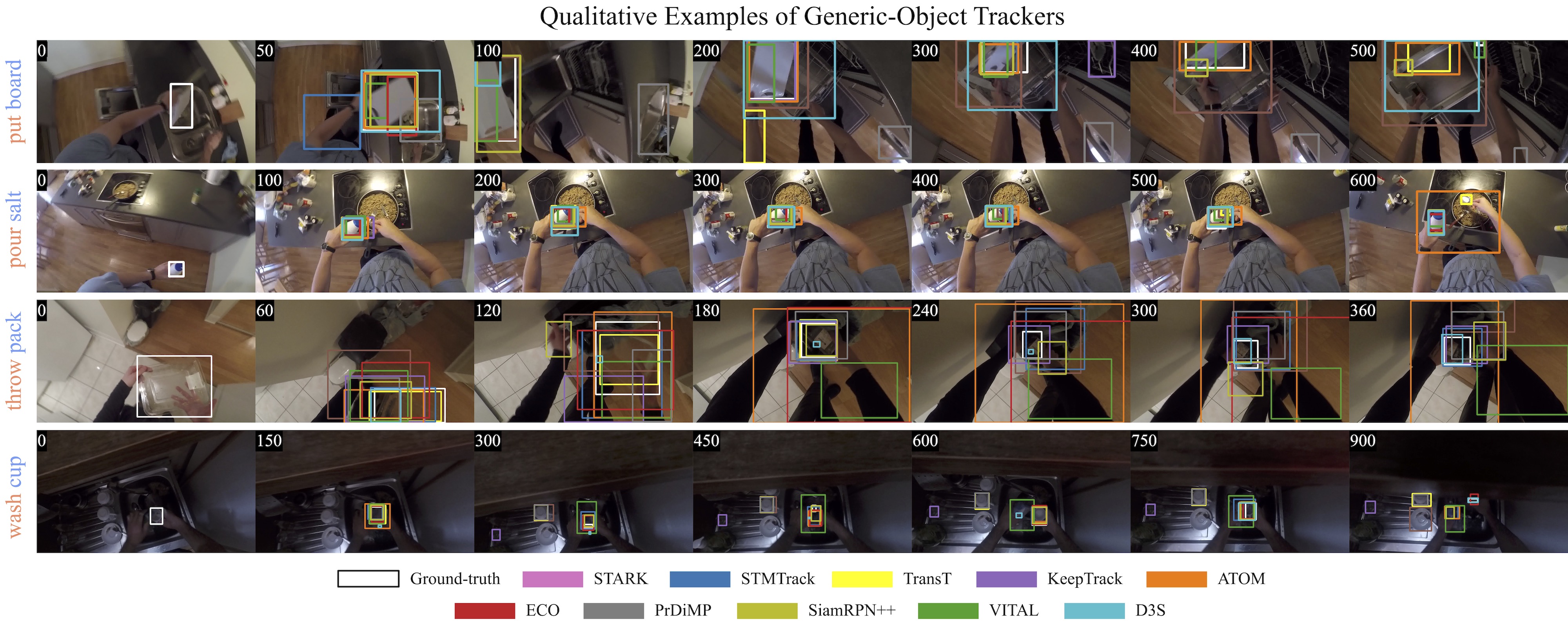}
\caption{Qualitative results of some of the generic object trackers benchmarked on the proposed \datasetname\ dataset.}
\label{fig:qualitativegot}
\end{figure*}

\section{Results}
\label{sec:resuluts}
\rev{In this section, we discuss the outcomes of our proposed study. For a better readability, in Figures and Tables we provide results for 20 of the 42 studied trackers. The results for all the trackers are given in the Figures and Tables of the supplementary document.}

\subsection{Performance of Generic Object Trackers}
\label{sec:resgot}
Figures \ref{fig:results} and \ref{fig:results42} report the performance of the generic object trackers on \datasetname\ using the OPE protocol, while Figures \ref{fig:resultsmse} and \ref{fig:resultsmse42} present the results achieved with the MSE protocol. Figure \ref{fig:qualitativegot} presents examples that qualitatively show the performance of some of the trackers.
Considering the results on a tracking approach basis, 
we have that trackers \AF{based on deep learning} (e.g. STARK, TransT, KeepTrack, LTMU, TrDiMP, ATOM, VITAL, ECO, Ocean) %
perform better than \AF{those based on}
hand-crafted features (e.g. BACF, MCCTH, DSST, KCF).
Among the first \AF{class of} trackers, the ones leveraging online adaptation mechanisms (e.g. STARK, STMTrack, KeepTrack, LTMU, TrDiMP, ATOM, VITAL, ECO, KYS, DiMP) are more accurate than \AF{the ones based on} single-shot instances (e.g. SiamGAT, Ocean, D3S, SiamBAN, SiamRPN++)
Trackers based on the transformer architecture \cite{Attention} (e.g. STARK, TransT, TrDiMP) hold the highest positions in the rankings of all the plots, suggesting that the representation learning and matching approach exploited by such trackers is suitable for better target-background discrimination in the FPV setting.
Indeed, the transformer-based matching operation between template and searching areas like the one implemented by STARK and TransT leads to a higher bounding box overlap on average (SS performance of Figure \ref{fig:results}(a)) and to a better centered bounding box (NPS performance of Figure \ref{fig:results}(b)).

Generally, the generalized success robustness plot in Figure \ref{fig:results}(c) and the GSR results of Figure \ref{fig:resultsmse} report different rankings of the trackers, showing that more spatially accurate trackers are not always able to maintain their accuracy for longer periods of time. Trackers that aim to build robust target models via online methods (e.g. STMTrack, ECO, TrDiMP, VITAL, MDNet, ATOM) result in better solutions for keeping longer temporal reference to objects. Particularly, the results achieved by STMTrack tell that a strategy based on memory networks building a highly dynamic representation of the template during tracking is beneficial to maintain a longer reference to the target.

By comparing the performance of the selected trackers with the results they achieve on standard benchmarks such as OTB-100 \cite{OTB}, as reported in Figure \ref{fig:opechange} \rev{of the supplementary document},
it can be noticed that the overall performance \AF{of all the trackers} is decreased across all measures \AF{when considering the FPV scenario}. 
\rev{Considering the extended usage of data driven approaches (e.g. deep learning) in visual tracking nowadays, we assessed the impact of leveraging large-scale FPV object localization data for training. \rev{In-depth discussion and results are provided in Section \ref{sec:fpvtrain} of the supplementary document. In short, some methodologies such as deep discriminative trackers \cite{DiMP} benefit from FPV-specific data, but the overall tracking performance still does not reach the quality that is observed in more common tracking benchmarks \cite{OTB,UAV123,NfS,VOT2019}. Other methodologies such as siamese network-based trackers \cite{SiamRPNpp} and transformer-based trackers \cite{Stark} are not able to exploit the context of  FPV from still FPV images. This weakness could be improved by yet-to-come large-scale FPV tracking datasets.}}
Overall, these outcomes demonstrate that, \rev{for the current availability of tracking data as well as the visual tracking knowledge in exploiting such}, the FPV setting poses new challenges to present trackers. 
\rev{It is worth mentioning that our achieved conclusions are consistent with the demonstrated performance drop of other object localization models (e.g. object detection) exploited between classical domains \cite{PASCAL,COCO} and FPV domains \cite{EK55}. }

\begin{table}[t]
\fontsize{7}{8}\selectfont
	\centering
	\caption{\rev{OPE and MSE performance of the baseline FPV-based tracking-by-detection methods TbyD-H and TbyD-F under different configurations. 
	} }
	\label{tab:tbyd}
	\setlength\tabcolsep{.1cm}
	\begin{tabular}{c | c | c  c  c | c  c  c }
		\toprule
		\multirow{2}{*}{Tracker} & \multirow{2}{*}{Version} & \multicolumn{3}{c|}{OPE}  & \multicolumn{3}{c}{MSE}  \\
                   & & SS & NPS & GSR  & SS & NPS  & GSR \\
		\midrule
		
		\multirow{3}{*}{TbyD-F} & SORT & 0.313 & 0.310 & 0.311 & 0.347 & 0.350 & 0.338 \\
		& IoU w prev box & 0.390 & 0.420 & 0.176 & 0.473 & 0.516 & 0.241 \\
		&  IoU w prev box + SORT & 0.390 & 0.425 & 0.192 & 0.476 & 0.521 & 0.264 \\
		
		\midrule
		
		\multirow{3}{*}{TbyD-H} & SORT & 0.237 & 0.213 & 0.222 & 0.264 & 0.252 & 0.241 \\
		&  IoU w prev box & 0.433 & 0.455 & 0.132 & 0.465 & 0.500 & 0.135 \\
		&  IoU w prev box + SORT & 0.432 & 0.457 & 0.137 & 0.465 & 0.502 & 0.142 \\

		\bottomrule		
\end{tabular}
\end{table}

\begin{table}[t]
\fontsize{7}{8}\selectfont
	\centering
	\caption{\rev{Performance of the proposed baseline FPV-trackers LTMU-H and LTMU-F applied over different trackers and under the OPE and MSE protocols used for the evaluation on \datasetname. 
	} }
	\label{tab:ltmufh}
	\setlength\tabcolsep{.18cm}
	\begin{tabular}{c | c | c  c  c | c  c  c }
		\toprule
		\multirow{2}{*}{Tracker} & \multirow{2}{*}{Version} & \multicolumn{3}{c|}{OPE}  & \multicolumn{3}{c}{MSE}  \\
                   & & SS & NPS & GSR  & SS & NPS  & GSR \\
		\midrule

		\multirow{3}{*}{DiMP-MU} & baseline & 0.411 & 0.432 & 0.320 & 0.445 & 0.469 & 0.342 \\
		& LTMU-F & 0.456 & 0.477 & 0.372 & 0.485 & 0.508 & 0.375 \\
		& LTMU-H & 0.461 & 0.486 & 0.376 & 0.495 & 0.517 & 0.380 \\
		
		\midrule
		
		\multirow{3}{*}{STMTrack} & baseline & 0.464 & 0.472 & 0.451 & 0.468 & 0.476 & 0.439 \\
		& LTMU-F & 0.461 & 0.471 & 0.408 & 0.471 & 0.481 & 0.411 \\
		& LTMU-H & 0.487 & 0.499 & 0.438 & 0.498 & 0.509 & 0.429 \\
		
		\midrule
		
		\multirow{3}{*}{STARK} & baseline & 0.492 & 0.504 & 0.395 & 0.530 & 0.546 & 0.413 \\
		& LTMU-F & 0.498 & 0.513 & 0.334 & 0.538 & 0.556 & 0.382 \\
		& LTMU-H & 0.505 & 0.520 & 0.370 & 0.543 & 0.561 & 0.382 \\

		\bottomrule		
\end{tabular}
\end{table}

\subsection{Performance of the FPV-specific Trackers}
\rev{The results achieved by the proposed TbyD-F and TbyD-H FPV-based tracking-by-detection baselines are compared with the generic object trackers in Figures \ref{fig:results}, \ref{fig:results42} and \ref{fig:resultsmse}, \ref{fig:resultsmse42}. As can be noticed, the baselines have competitive results with the best trackers in the SS and NPS metrics, but they struggle in the GSR. This means that they are not able to maintain reference to the objects even though the other scores suggest they provide spatially accurate localizations. By comparing TbyD-F with TbyD-H, we observe that the second is better in an OPE-like execution scenario, while the first achieves higher scores in the MSE experiments. Table \ref{tab:tbyd} reports the performance of such two trackers with other strategies (details are given in Section \ref{sec:tbydsupp} of the supplementary document) that implement target association on top of object detection \cite{SORT,TAO}. A simple application of SORT \cite{SORT} does not work as well as demonstrated in other domains \cite{TAO}, and applying such method in combination with the strategy described in Section \ref{sec:tbyd} brings little benefit.}

\rev{Figures \ref{fig:results}, \ref{fig:results42} and \ref{fig:resultsmse}, \ref{fig:resultsmse42} also show the performances of the other FPV baselines LTMU-F and LTMU-H  in comparison with the different trackers.} 
\rev{In both the OPE and MSE experiments, the proposed trackers achieve the top spots in the SS and NPS rankings, while they lose some performance in the GSR score.}
\rev{Table \ref{tab:ltmufh} shows the performance gain in applying the LTMU-F/H scheme over different generic object trackers \cite{LTMU,STMTrack,Stark}. Overall, both LTMU-F and LTMU-H increase the SS and NPS metrics of the underlying tracker, with the second presenting a generally larger improvement. In the versions with STARK and STMTrack, the GSR scores are decreased. However, looking at the DiMP-MU version (as used in \cite{LTMU}) we see that the performance is improved by a good margin in all the metrics, including the GSR. Considering that such an underlying tracker uses a MetaUpdater \cite{LTMU} to better assess the consistency of the tracker in triggering re-detection and model update, we hypothesize that such a module could bring benefit to the other versions if properly customized to. }
Figure \ref{fig:qualitativefpv} \rev{of the supplementary document} presents some qualitative examples of the performance of the LTMU-F/H trackers in contrast to the baseline one. 
Overall, the message to take from these outcomes is that adapting a state-of-the-art method with FPV-specific components allows to increase the tracking performance. 
\rev{Combining hand and object tracking, as the baseline LTMU-H na\"ively does, results a promising direction.} 
We hence expect significant performance improvements to be achievable by a tracker accurately designed to exploit FPV-specific cues such as the characteristics of the interaction between the target and the camera wearer.

\subsection{\rev{Initialization by an Object Detector}}
\begin{figure}[t]%
\centering
\includegraphics[width=\columnwidth]{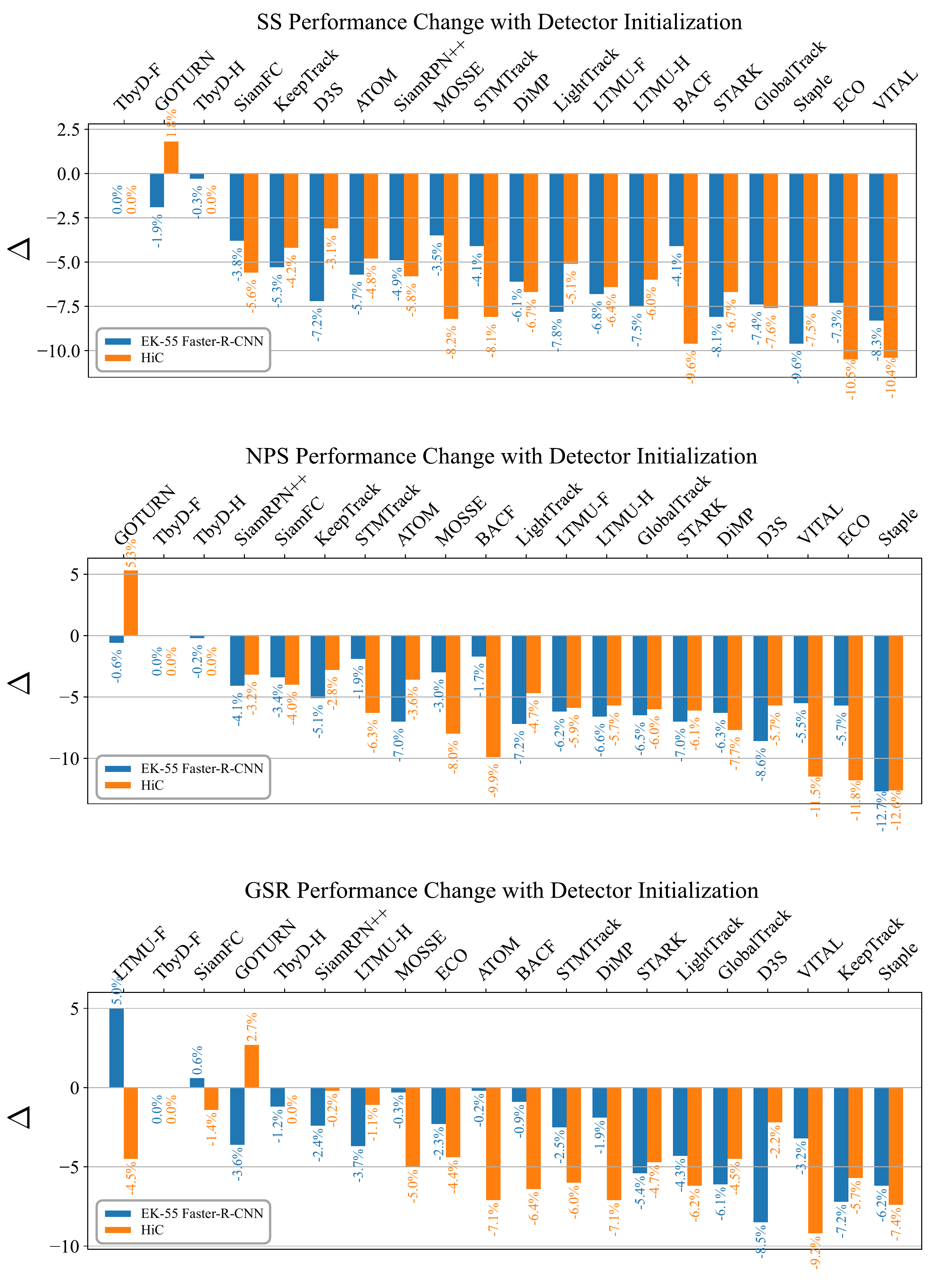}
\caption{\rev{Results of the OPE-D experiment in which the bounding box for initialization is given either by the EK-55 trained Faster-R-CNN \cite{EK55} or the HiC detector \cite{Shan2020}. The performance change (in percentage) for 20 of the selected trackers with respect with the ground-truth initialization is reported for the SS, NPS, and GSR metrics. The trackers are ordered by the average performance change.}}
\label{fig:oped}
\end{figure}

\rev{Figures \ref{fig:oped}, \ref{fig:oped42} report the SS, NPS, GSR performance change when the EK-55 Faster-R-CNN \cite{EK55} or the HiC \cite{Shan2020} detection bounding box  is used to initialize the trackers. 
In general, such a process causes a drop in the tracking performance. This can be explained by the noise in the position and scale of the initial target state that consequently affects the constructions of the models that are used for tracking during the video \cite{OTB}. 
By computing the average delta across the trackers for each of the metrics, we obtain that Faster-R-CNN causes SS, NPS, GSR drops of -5.3\%, -5.1\%, -3.1\%. HiC leads to slightly larger drops of -5.9\%, -5.7\%, -4.4\%. It is worth mentioning that Faster-R-CNN provided 149 valid detections out of 150 with an average delay of 14 frames from the start of the sequence, while HiC gave 146 valid detections with a delay of 28 frames. Hence, HiC is a  weaker object detector. Overall, we consider the average performance drop quite limited, thus making the trackers usable even in cases of noisy initialization. TbyD-F/H are among the trackers losing less accuracy, but despite this their performance does not surpass trackers more susceptible to noise, such as LTMU-F/H, STARK, TransT. Indeed, when initialized by Faster-R-CNN, TbyD-H achieves SS 0.440, while LTMU-H, STARK, and TransT, achieve SS 0.478, \linebreak 0.470, 0.466, respectively.}

\begin{figure}[!ht]%
\centering
\includegraphics[width=\columnwidth]{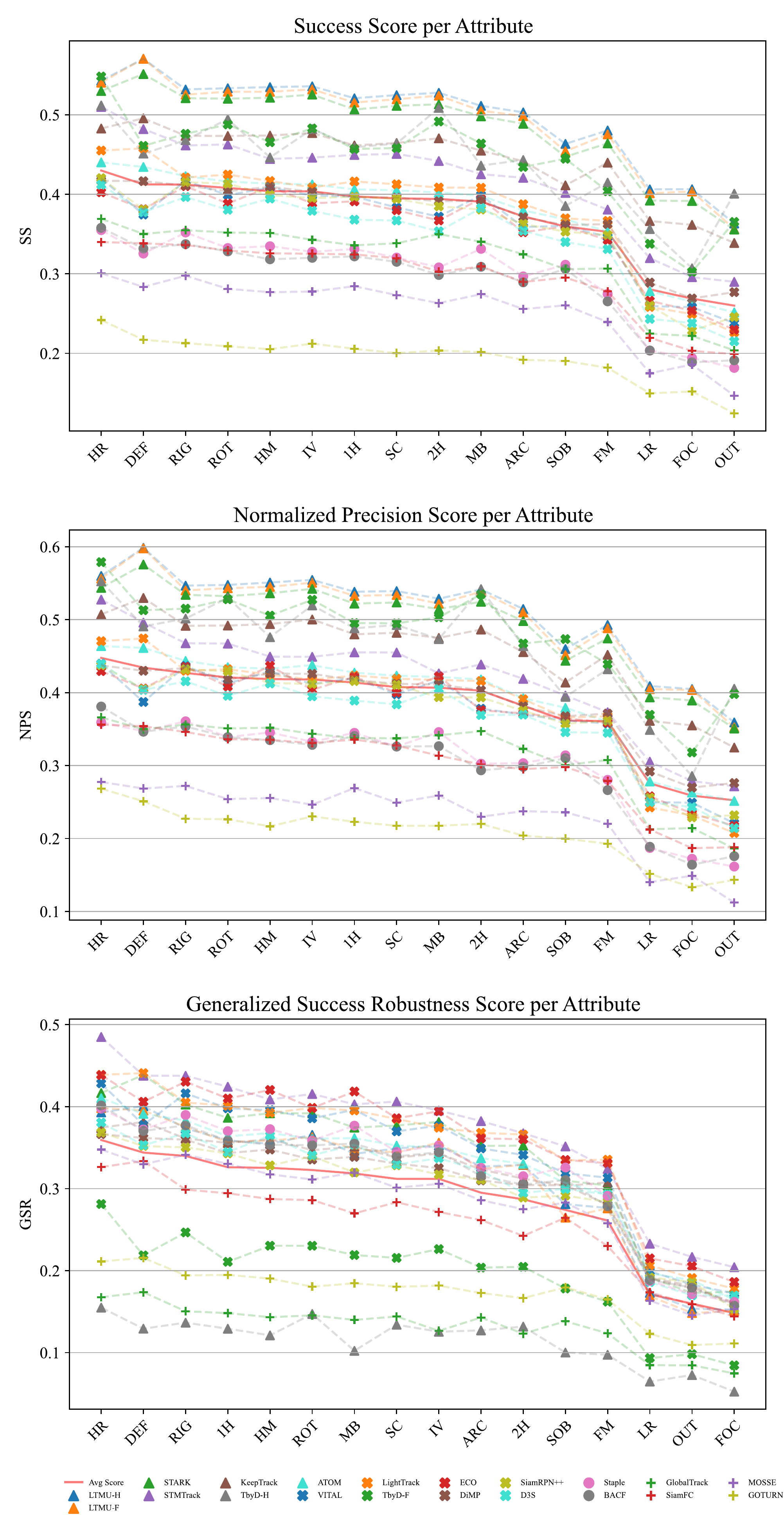}
\caption{SS, NPS, and GSR performance achieved under the MSE protocol of \rev{20 the 42} selected trackers with respect to the sequence attributes available in \datasetname. (The results for the POC attribute are not reported because this attribute is present in every sequence). The red plain line highlights the average tracker performance.}
\label{fig:resattributes}
\end{figure}

\begin{figure}[!ht]%
\centering
\includegraphics[width=\columnwidth]{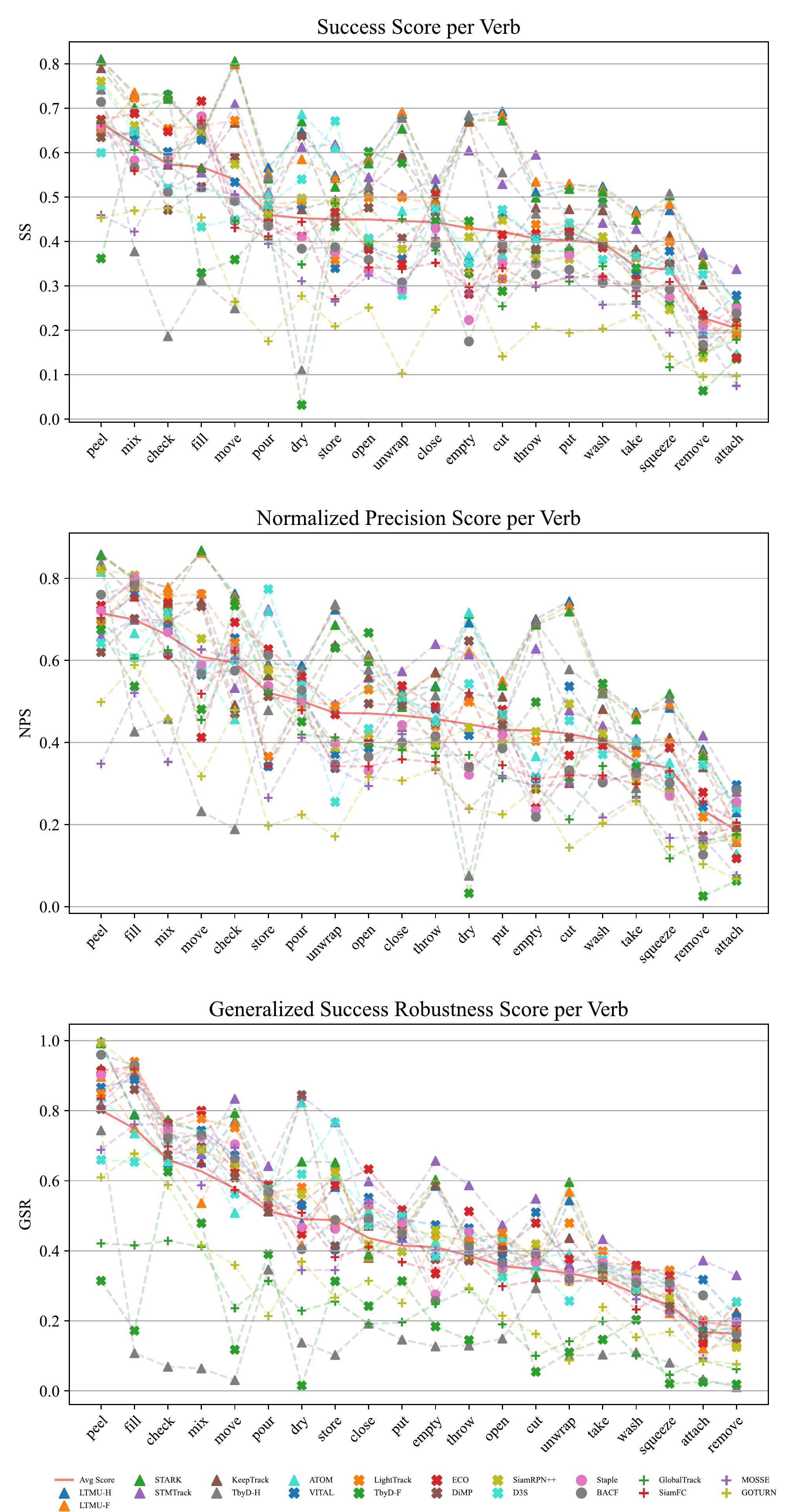}
\caption{SS, NPS, and GSR performance achieved under the MSE protocol of \rev{20 of the 42} selected trackers with respect to the action verbs performed by the camera wearer and available in \datasetname. The red plain line highlights the average performance.}
\label{fig:resverbs}
\end{figure}

\begin{figure}[!ht]%
\centering
\includegraphics[width=\columnwidth]{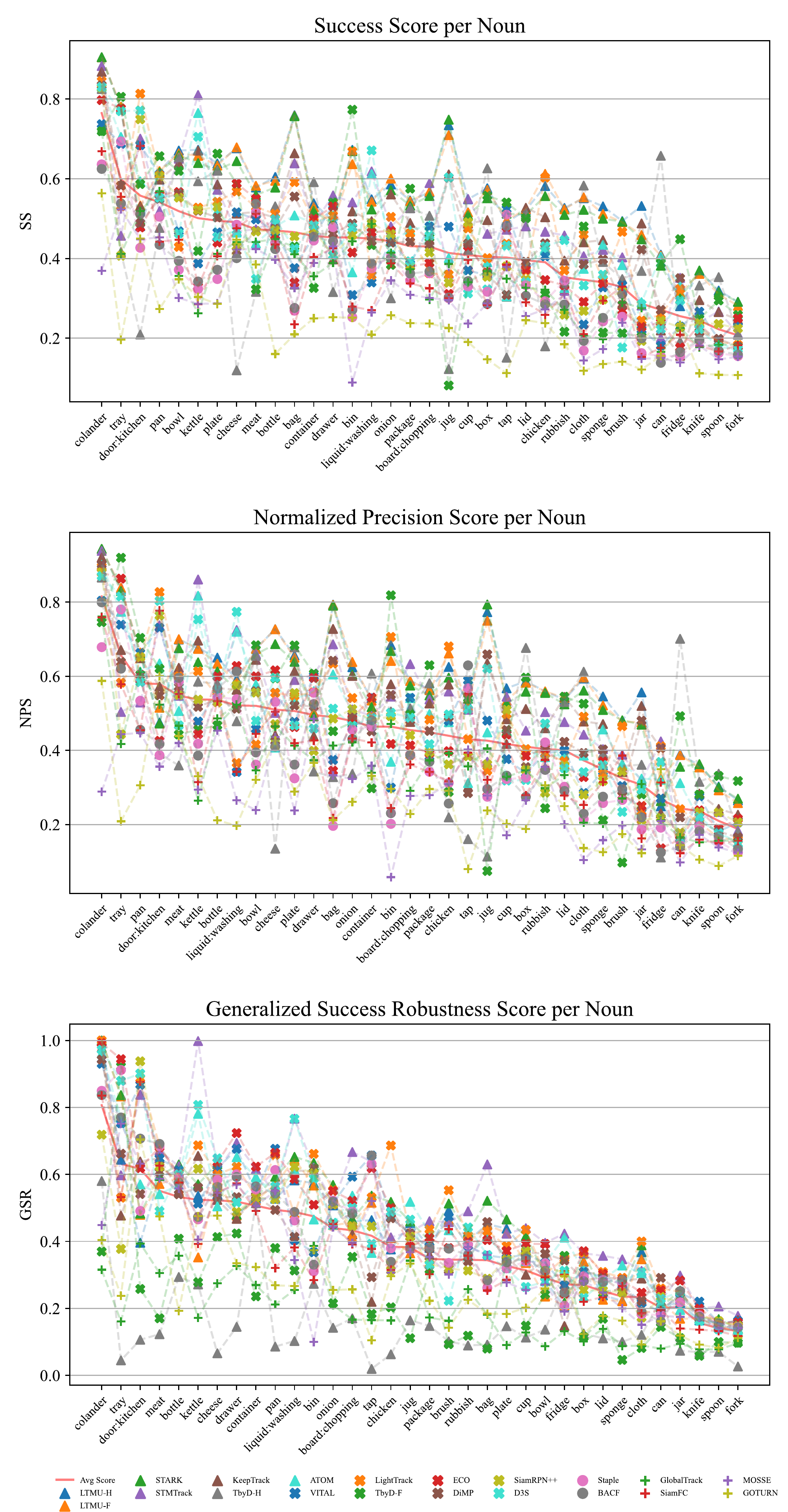}
\caption{SS, NPS, and GSR performance achieved under the MSE protocol of \rev{20 the 42} selected trackers with respect to the target noun categories available in \datasetname. The red plain line highlights the average tracker performance.}
\label{fig:resnouns}
\end{figure}

\subsection{Attribute Analysis}
Figure \ref{fig:resattributes} \AF{reports} the SS, NPS, and GSR scores, computed with the MSE protocol, of the \rev{20 representative} trackers with respect to the 
attributes introduced in Table \ref{tab:attrdesc}.
We do not report results for the POC \AF{attribute} as it is present in every sequence, as shown in Figure \ref{fig:attrdistributions}~(a).
It stands out clearly that full occlusion~(FOC), out of view~(OUT) and the small size of targets~(LR) \AF{are} the most difficult situations for all the trackers. The fast motion of targets~(FM) and the presence of similar objects~(SOB) are also critical factors that cause drops in performance. 
Rotations~(ROT) and the illumination variation~(IV) are better addressed by the trackers. The algorithms also do not demonstrate significant behavior changes between the tracking of rigid or deformable objects. 
With respect to the new \AF{4} sequence attributes related to FPV, the results report that tracking objects held with two hands~(2H) is more difficult than tracking objects held with a single hand~(1H). This is because the manipulation of the target by two hands generates situations in which the occlusions are more extended over the object's appearance.
Trackers are instead quite robust to the \AF{head motion}~(HM), which influences the camera movements,
\AF{and seem to cope well with objects appearing in larger sizes~(HR).}

In terms of algorithmic principles, we have that STARK has better SS results over the second-best \rev{generic object tracker}, TransT, across all the conditions described by the attributes except for the case of deformable objects~(DEF) and the presence of similar objects~(SOB). In the latter situations, the performance of the two trackers is around the same. For the NPS, STARK results better than TransT in general, even though the gap between them is reduced. \linebreak TransT outputs better centered bounding boxes in the DEF and SOB conditions.
Considering the GSR measure, we observe that STMTrack results in the best methodology across most of the attributes. The improvement over the other solutions is particularly significant in the presence of the challenging conditions of small objects~(LR), \linebreak target out-of-view~(OUT), and full occlusion~(FOC). STMTrack exhibits also a much better score with objects appearing in large size~(HR). The ECO tracker instead provides longer references to targets in the case of head motion~(HM), motion blur~(MB), and fast motion~(FM). 
With respect to the introduced FPV trackers, we have that the performance of STARK is improved by LTMU-H and LTMU-F overall. 
\rev{The TbyD-H tracker has a particularly higher SS and NPS performance in out of view conditions~(OUT), suggesting a capability in finding again the targets after the re-apperance in the scene.}
These outcomes tell that the introduced FPV-specific components are particularly helpful in the circumstances that affect the trackers the most.

\subsection{Action Analysis}
The plot in Figure \ref{fig:resverbs} reports
the MSE protocol results of SS, NPS, and GSR with respect to the action verb labels associated to the actions performed by the camera wearer in each video sequence. %
\rev{We think that the results presented in the following can give cues about the exploitation of trackers for action recognition tasks.}
 In general, we observe that the actions \AF{mainly} causing a spatial \AF{displacement} 
of the target (e.g. ``move'', ``store'', ``check'') have less impact on the performance of the trackers.
Instead, actions that change the state, shape, or aspect ratio of the target object (e.g. ``remove'', ``squeeze'', ``cut'', ``attach'') generate harder tracking scenarios.
\AF{Also the sequences characterized by the ``wash'' action verb lead trackers to poor performance.}
Indeed, such an action makes the object harder to track
because of the many occlusions caused by the persistent and severe manipulation washing involves.
It can be noted from the plots that no tracker prevails overall, but LTMU-F/H, STARK, and TransT occupy the top stops especially in the plots relative to SS and NPS. 
In general, the performance of the trackers varies much across the different actions showing that various approaches are suitable to track under the different conditions generated.

The plots in Figure \ref{fig:resnouns} presents the performance scores of the trackers with respect to the target noun labels, i.e. the categories of target object.
Rigid, regular-sized objects such as ``pan'', ``kettle'', ``bowl'', ``plate'', and ``bottle'' are among the ones associated with higher average SS greater or around 0.5, but some of them (e.g. ``plate'' and ``bottle'') lead to lower GSR scores meaning that trackers provide a spatially accurate but short temporal reference to such kind of objects.
In contrast, other rigid objects such as ``knife'', ``spoon'', ``fork'' and ``can'' are more difficult
to track from the point of view of all the considered measures (the scores are around 0.3 or lower). This is probably due to the particularly thin shape of these objects and the light reflectance they are \AF{easily} subject to.  
\AF{Deformable objects such as ``sponge'', ``onion'',  ``cloth'' and ``rubbish'' are in general also difficult to track.}

\subsection{\rev{Hand-Object Interaction Evaluation}}
Tables \ref{tab:handsobjhic}, \ref{tab:handsobjhicfull}  present the results of \rev{the evaluation of the HOI task described in Section \ref{sec:evalfpv} in relation to the considered trackers}.
Despite we are showing that FPV introduces challenges for current trackers, with this experiment we want to assess whether they can be still exploited in the FPV domain to obtain information about the objects' locations and movements in the scene~\cite{Wang2020,Furnari2017,RULSTMpami,sener2020temporal,Shan2020}.
\rev{The results given in the first column of the table report the Recall of the proposed video-based HOI  detection pipeline in which each tracker is included. The values in the brackets of the second column report the SS, NPS, and GSR results achieved by the tracker run in an OPE-like fashion on the same sub-sequences on which the pipeline is executed.}
\rev{It can be noticed how the performance difference between the trackers is reduced with respect to what showed in Figures \ref{fig:results}, \ref{fig:results42}. This demonstrates that when deployed for HOI, the different tracking methodologies lead to an overall similar pipeline.} Particularly, it results that STARK is a better suited methodology for tracking objects starting from an initialization given by an object detection algorithm in this context.
By comparing the Recall with the tracker performance scores (SS, NPS, GSR), it can be noted that there is a correlation between the first and the SS, since the ranking of the trackers according to the first measure is very similar to the one of the second measure.

\rev{In Table \ref{tab:handsobjgtfull} of the supplementary document, the results of an oracle-based solution that gives the optimal bounding box for the interacted object at the first frame of HOI are presented.
The first thing that stands out is the performance gap with respect to what reported in Tables \ref{tab:handsobjhic}, \ref{tab:handsobjhicfull}. This is due to the performance of HiC  which struggles to find a valid HOI detection in the proposed video-based pipeline. This issue delays the initialization of the tracker making the overall pipeline not detecting and localizing the HOI in many frames. }
These outcomes show that, if initialized with a proper bounding box for the object involved in the interaction, the trackers are able to maintain the spatial and temporal reference to such an object for all the interaction period with promising accuracy. Indeed, the Recall values achieved by the proposed HOI system with LTMU-H reaches 0.754. 
It is also worth observing that the SS, NPS, GSR scores achieved in this experiment reflect the performance achieved by the trackers with the OPE protocol on the full sequences of \datasetname, as reported in Figures \ref{fig:results}, \ref{fig:results42}. These results demonstrate that the evaluation of the trackers' performance on the original sequences of \datasetname\ can lead to conclusions about the behavior of the trackers in particular FPV application scenarios.
Furthermore, the reader might wonder why there is such a large absolute difference in the values of the SS, NPS, and GSR present in Table \ref{tab:handsobjgtfull} and those in the brackets of Figure \ref{fig:results42}. This can be explained by the fact that in the considered HOI evaluation the lengths of the video sequences are very short (the average length is of \rev{81} frames). 
In contrast, the average length of the full video sequences present in \datasetname\ is 649 frames, which is much higher than the previously discussed number. 
Such a shorter duration of the videos simplifies the job of the trackers since the variations of the target object and the scene are less significant in these conditions rather than in longer sequences. A justification to this explanation is also given by the GSR results of Figures \ref{fig:results}, \ref{fig:results42}. For example, on such measure, STARK achieves 0.395 which means that such an algorithm tracks successfully until the 39.5\% of a sequence length. In number of frames, such a fraction is 256 on average. This value is much higher than the length of the sub-sequences and explains why the performance of STARK is so successful in the context of this FPV application.
Furthermore, in the oracle-based HOI experiment we observe that the ranking of the trackers slightly changes. Trackers that reached lower spots in this experimental setting (e.g. TbyD-H, LTMU, Ocean, D3S), in the HiC-based pipeline compete in making the HOI system more accurate (i.e. they increase the Recall).
Considering that in the latter situation the initialization box is not as accurate as the ground-truth, such an outcome additionally confirms that the different trackers are subject in a different manner to the initialization noise.

\begin{table}[!t]
\fontsize{8}{9}\selectfont
	\centering
	\caption{\rev{Results of the experiment in which 20 of the considered trackers are evaluated by the Recall of an FPV HOI detection pipeline where trackers are used as localization method for the object involved in the interaction. The first column presents the results of the proposed system in which each tracker is initialized with the bounding box given by HiC in its first valid HOI detection. The last column reports the SS, NPS, and GSR results achieved by each tracker with the OPE protocol on the sub-sequences yielded by the HOI labels. Best results, per measure, are highlighted in \tblbest{gold}, second-best in \tblsecondbest{silver}, third-best in \tblthirdbest{bronze}.}}
	\label{tab:handsobjhic}
	\setlength\tabcolsep{.5cm}
	\rowcolors{1}{tblrowcolor1}{tblrowcolor2}
	\begin{tabular}{l | c c }
		\toprule
		Tracker & Recall & (SS, NPS, GSR) \\
		
		\midrule
		
		STARK & \tblbest{0.248} & (\tblbest{0.211}, \tblthirdbest{0.221}, \tblsecondbest{0.222}) \\
		LTMU-H & \tblsecondbest{0.246} & (\tblsecondbest{0.210}, \tblsecondbest{0.222}, 0.217) \\ 
		LTMU-F & \tblthirdbest{0.245} & (\tblsecondbest{0.210}, \tblthirdbest{0.221}, 0.216) \\
		TbyD-H & 0.238 & (\tblthirdbest{0.205}, \tblbest{0.223}, 0.163) \\
		LightTrack & 0.233 & (0.197, 0.212, \tblbest{0.228}) \\
		KeepTrack & 0.232 & (0.201, 0.214, 0.212)  \\
		SiamRPN++ & 0.227 & (0.191, 0.206, 0.209) \\ 
		TbyD-F & 0.220 & (0.184, 0.202, 0.179) \\
		STMTrack & 0.216 & (0.196, 0.202, \tblthirdbest{0.219})   \\

		D3S & 0.211 & (0.187, 0.199, 0.208) \\ 
		ECO & 0.211 & (0.181, 0.196, 0.217) \\ 
		DiMP & 0.210 & (0.186, 0.198, 0.211) \\
		ATOM & 0.207 & (0.186, 0.198, 0.213) \\ 
		VITAL & 0.198 & (0.178, 0.192, 0.213) \\ 
		SiamFC & 0.195 & (0.171, 0.180, 0.195) \\ 
		GlobalTrack & 0.195 & (0.170, 0.180, 0.144) \\ 
		BACF & 0.188 & (0.170, 0.189, 0.206) \\
		Staple & 0.182 & (0.164, 0.179, 0.204) \\  
		MOSSE & 0.158 & (0.151, 0.154, 0.188) \\ 
		GOTURN & 0.139 & (0.138, 0.147, 0.162) \\

		\bottomrule	
		
\end{tabular}
\end{table}

\subsection{Contribution of Trackers to FPV Tasks}
To understand if the employment of trackers brings advantages with respect to the more standard object localization solutions used in FPV \cite{Shan2020,EK55}, we compared the Recall results of the trackers presented in Table \ref{tab:handsobjhic} with the Recall results of the original hand-object interaction detector HiC \cite{Shan2020} which processes the frames independently. This solution achieves a Recall of \rev{0.113} which results very low when compared to the \rev{0.248, 0.246, and 0.245} achieved by the pipelines exploiting \rev{STARK, LTMU-H, TransT,} respectively.

\rev{In addition, we compared the performance the EK-55-trained Faster R-CNN~\cite{EK55} and HiC \cite{Shan2020} when used as pure object detectors (not exploiting temporal information for \linebreak tracking as in the TbyD-F/H baselines)}. 
\rev{In this case, for Faster-R-CNN, at every frame, we consider as output the bounding box having the highest score associated to the category of the target object in the video, while for HiC we just take the object bounding box having the largest score (HiC provides class-agnostic object detections).}
On the sequences of \datasetname\, the first solution achieves an OPE-based SS, NPS, and GSR of 0.323, 0.369, 0.044 respectively, and runs at 1 FPS, \rev{while the second reaches SS 0.411, NPS 0.438, GSR 0.007, at 8 FPS}. 
\rev{Comparing these results with those of the TbyD-F/H baselines, we see the advantage of performing tracking, since all the metric scores are improved. Moreover, if we} compare the detectors' results with the ones presented in the overall study, we clearly notice that trackers, \rev{even when} initialized by a detection module, can deliver faster, more accurate, and much temporally longer object localization than detectors.

Overall, these outcomes demonstrate that visual object trackers can bring benefits to FPV application pipelines. In addition to the ability of maintaining reference to specific object instances, the advantages of tracking are achieved in terms of better object localization and efficiency. We hence expect that trackers will likely gain more importance in FPV as new methodologies explicitly considering the first person point of view are investigated.

\section{Conclusions}
In this paper, we \AF{proposed} the first systematic evaluation of visual object tracking in first person vision (FPV). The analysis \AF{has been} conducted with standard and novel measures on the newly introduced \datasetname\ benchmark, \AF{which} contains 150 video sequences extracted from the EK \cite{EK55,EK100} FPV dataset. \datasetname\ has been densely annotated with 97K bounding-boxes, 17 sequence attributes, 20 action verb attributes, and 34 target object attributes, as well as with \rev{167K} spatial annotations for the camera wearer's hands and \rev{78K} states of interaction with the target object.
The performance of 38 state-of-the-art generic object visual trackers and \rev{four} baseline FPV trackers was analysed extensively on the proposed dataset. 
The investigation has conducted to the following conclusions. The performance of all the benchmarked trackers is decreased when compared with the respective accuracy on other popular visual object tracking benchmarks. This is explained by the different nature of images and the particular characteristics introduced by FPV which offer new and challenging conditions \rev{for the current knowledge in the visual tracking domain and the lack of tracking-specific FPV data}. The analysis revealed that deep learning-based trackers employing online adaptation techniques achieve better performance than the trackers based on siamese neural networks or on handcrafted features. \rev{Among the different methodologies based on this approach, the transformer-based worked the best and hence is a promising future direction. This exploration could involve the curation of large-scale diverse tracking-specific data.} 
The introduction of FPV-specific object localization modules, \rev{such as HOI models}, in a tracking pipeline increased its performance, demonstrating that particular cues about the domain influence the tracking accuracy. \rev{These results highlighted the potential direction of joint hand-object tracking, and we expect successful methodologies to take into account also cues about the camera wearer's surroundings.}
The performance of the trackers was then studied in relation to specific attributes characterising the visual appearance of the target and the scene. It turned out that the most challenging factors for trackers are the target's out of view, its full occlusions, its low resolution, as well the presence of similar objects or of fast motion in the scene. Trackers were also analyzed based on the action performed by the camera wearer as well as the object category the target belongs to. It resulted that actions causing the change of state, shape, or aspect ratio of the target affected the trackers more than the actions causing only spatial changes. \rev{We think that trackers incorporating semantic information about the person's action could be an interesting direction of investigation.}
We observed that rigid thin-shaped objects are among the hardest ones to track. Finally, we evaluated the trackers in the context of the FPV-specific application of video-based hand-object interaction detection. We included each tracker in a pipeline to tackle such a problem, and evaluated the performance of the system to quantify the tracker's contribution. We observed that the trackers demonstrate a behavior that is consistent with their overall performance on the sequences of \datasetname. Even though FPV introduced challenging factors for trackers, the results in such a specific task demonstrated that current trackers can be used successfully if the video sequences in which tracking is required are not too long. We also demonstrated that trackers bring advantages in terms of object referral and localization, and efficiency, over object detection.
\rev{We think that an effective and efficient integration of tracking methodologies with those of FPV downstream applications is a relevant problem to study.}
In conclusion, we believe that there is potential in improving FPV pipelines by employing visual trackers as well as there is room for the improvement of the performance of visual object trackers in this new domain.

\paragraph{Acknowledgements} Research at the University of Udine has been supported by the ACHIEVE-ITN H2020 project. Research at the University of Catania has been supported by MIUR AIM - Attrazione e Mobilita Internazionale Linea 1 - AIM1893589 - CUP: E64118002540007.

\clearpage

\twocolumn[{%
\centering
\vspace{1em}
{\Large  ``Visual Object Tracking in First Person Vision''} \\
\vspace{.5em}
{\large  Supplementary Document} \\
\vspace{1em}
Matteo Dunnhofer, Antonino Furnari, Giovanni Maria Farinella, Christian Micheloni \\
Corresponding author e-mail: \href{matteo.dunnhofer@uniud.it}{mailto:matteo.dunnhofer@uniud.it}
\vspace{2.5em}
}]

\section{The TREK-150 Benchmark}

\subsection{Further Motivations and Details}
\label{sec:datasetdiff}
In this section, we provide further motivations and details behind the construction of the \datasetname\ dataset.

\paragraph{Frame Rate} 
The video sequences included in \datasetname\ have the frame rate of 60 FPS inherited from EK. According to the authors \cite{EK55,EK100}, EK has been acquired with such a  frame rate because of the proximity of the camera point of view and the main scene (i.e. manipulated objects) which causes very fast motion, and heavy motion blur due to the camera wearer movements (especially when he/she moves the head). We empirically checked the amount of fast motion by assessing the average normalized motion happening between the bounding boxes of consecutive frames that include such a condition. The motion has been quantified as the distance between the center of two consecutive ground-truth bounding boxes normalized by the frame size. Considering a subsampled version of \datasetname\ at 30 FPS, such a value achieves 0.075. This is higher than the values present in other tracking benchmarks such as the 0.068 of OTB-100, the 0.033 of UAV123, or the 0.049 of the 30 FPS-version of NfS. These comparisons demonstrate that the FPV scenario effectively includes challenging conditions due to the faster motion of the targets/scene. Considering the original frame rate of 60 FPS, the fast motion quantity of \datasetname\ is reduced to 0.062, which is comparable to the values obtained in other tracking benchmarks.

\begin{figure}[t]%
\centering
\includegraphics[width=\columnwidth]{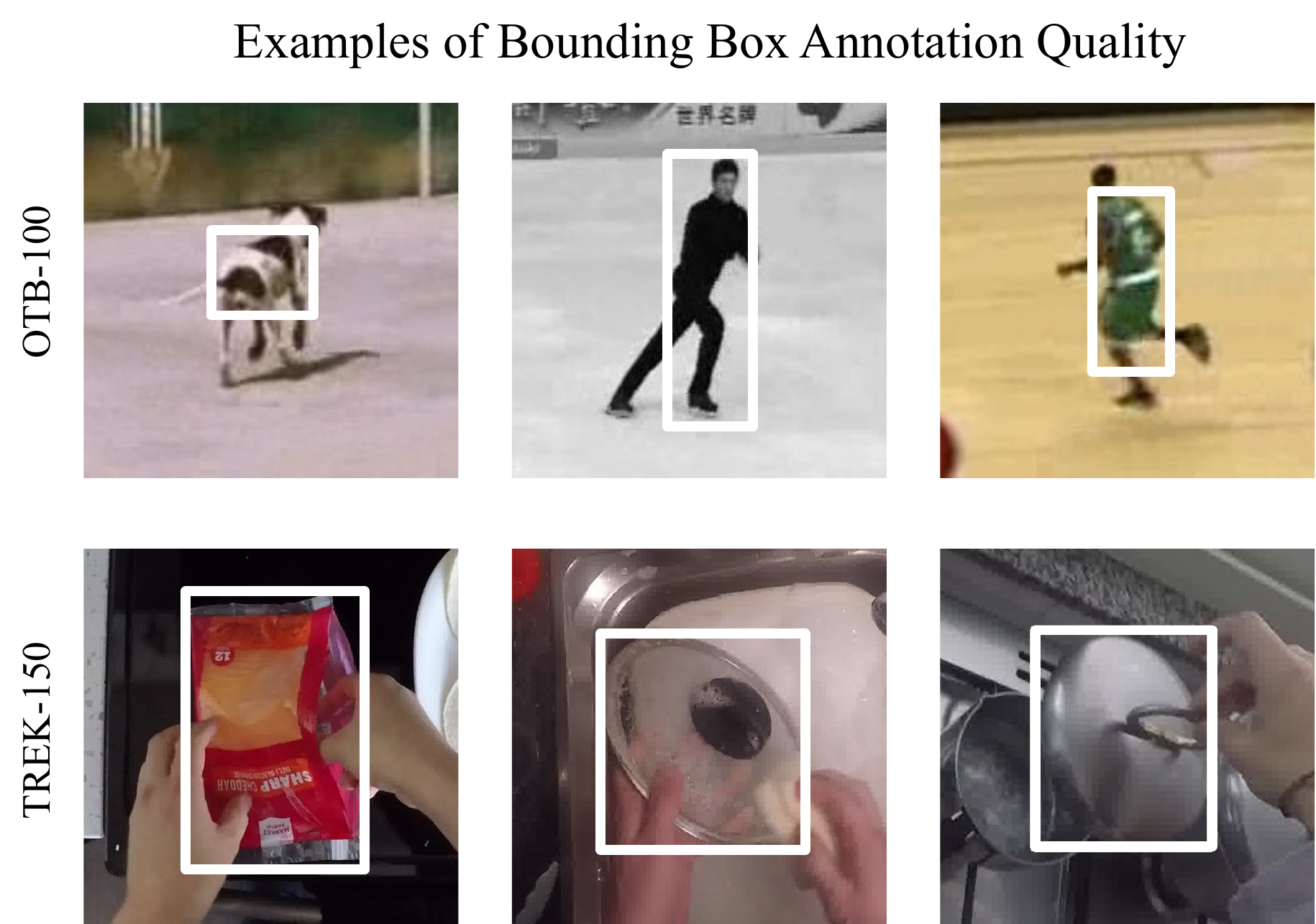}
\caption{Examples of the quality of the bounding box annotations contained in \datasetname\ in comparison with the ones available in the popular OTB-100 benchmark. \datasetname\ provides careful and high-quality annotations that tightly enclose all the target objects.}
\label{fig:annoquality}
\end{figure}

\begin{figure*}[t]%
\centering
\includegraphics[width=.7\linewidth]{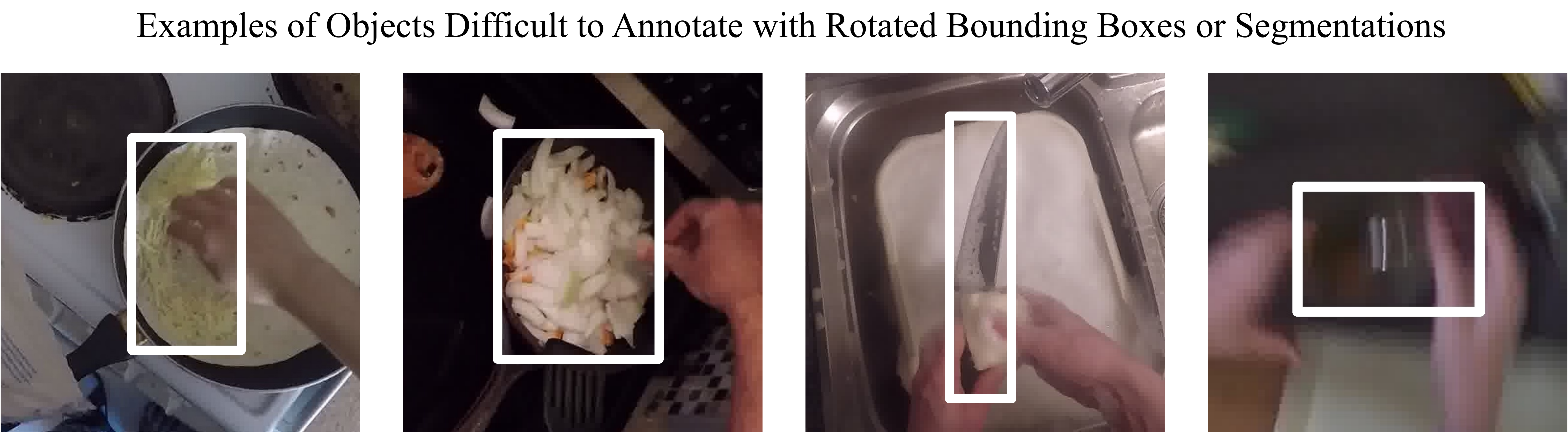}
\caption{Examples of target objects contained in \datasetname\ that are difficult to represent with more sophisticated representations (e.g. rotated bounding box or segmentation mask). The first two images from the left show objects such as ``cheese'' and ``onion'' which prevent the determination of the angle for an oriented bounding-box, or an accurate segmentation mask. The last two images present objects which prevent a consistent definition of a segmentation mask.}
\label{fig:hardex}
\end{figure*}

\paragraph{Sophisticated Target Representations}
We would like to point out the difficulty that the FPV setting poses on the creation of more sophisticated annotations for the object categories appearing commonly in FPV scenarios. Figure \ref{fig:hardex} shows some examples of these. The first two images from the left show the objects “cheese” and “onion” (these are considered as single objects according to the EK-55 annotations \cite{EK55}) which prevent the determination of the angle for an oriented bounding box, or an even accurate segmentation mask due to their spatial sparsity. The two images on the right present objects for which providing a segmentation is very ambiguous. Indeed, most of the pixels in the image area of the knife (third image) belong actually to foam, while the heavy motion blur happening on the object of the fourth image (where the target is a bottle) prevents the definition of the actual pixels belonging to the object. In all these scenarios, axis-aligned bounding boxes result in robust target representations that provide a consistent delineation of the object. 
Hence, to make the annotations consistent across the whole dataset, we employed such representations for \datasetname.

\paragraph{Sequence Annotations}
To study the performance of trackers under different aspects, the sequences of \datasetname\ have been associated with one or more of 17 attributes that indicate the visual variability of the target in the sequence (see Table \ref{tab:attrdesc} of the main paper for the details). The extended usage of this practice \cite{OTB,UAV123,NfS,NUSPRO,TrackingNet,LaSOTijcv,GOT10k} showed how this kind of labeling is sufficient to estimate the trackers' performance on particular scenarios. 
We therefore followed such an approach to associate labels on \datasetname's videos. 
However, we argue that, by using this labeling setting, attention must be paid to how trackers are evaluated. The standard OPE protocol, which has been generally used to perform such evaluations, could lead to less accurate estimates. For example, it could happen that a tracker would fail for some event described by an attribute (e.g. FOC) in the first frames of a video, but that the sequence also contains some other event (e.g. MB) in the end. With the score averaging procedure defined by the OPE protocol, the low results achieved due to the first event would set low scores also for the second event, while the tracker failed just for the first one. Therefore, the performance estimate for the second attribute would not be realistic. 
We believe that a reasonable option is to use a more robust evaluation protocol such as the multi-start evaluation (MSE).
Thanks to its points of initialization which generate multiple diverse sub-sequences, this protocol allows a tracker to better cover all the possible situations happening along the videos, both forward and backward in time. All the results achieved on the sub-sequences are then averaged to obtain the overall scores on a sequence. We think the scores computed in this way to be more robust and accurate estimates of the real performance of the trackers. Hence, in this work, we follow such an approach to evaluate trackers over sequence attributes.

Even though we provide per-frame labels to describe the interaction happening between the camera wearer and the target object (LHI, RHI, BHI labels), we considered the sequence attributes 1H and 2H in order to have a direct and more consistent assessment of the impact of the conditions they indicate in relation to the other attributes associated to sequences.

\begin{figure*}[!ht]%
\centering
\includegraphics[width=\linewidth]{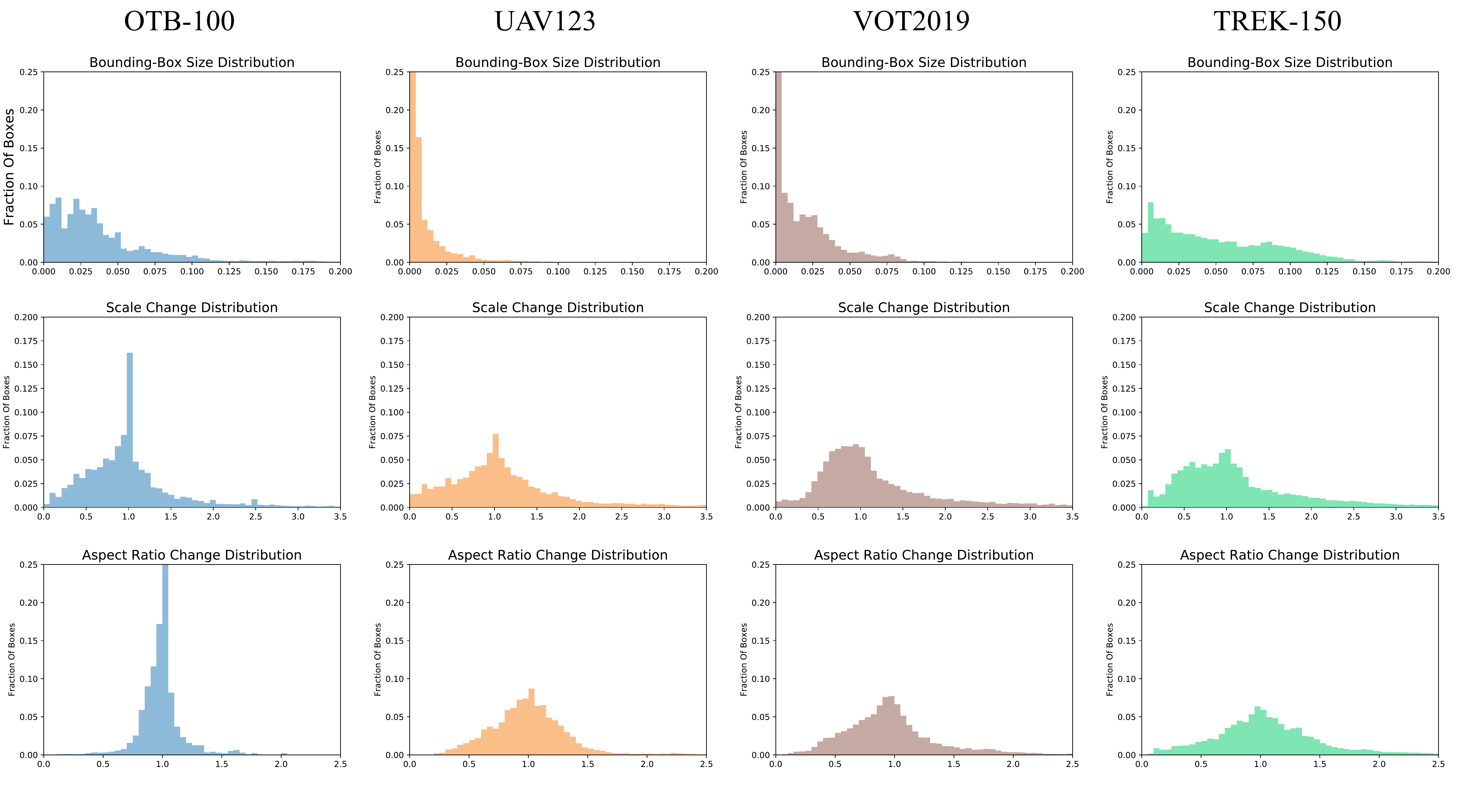}
\caption{Comparison between \datasetname\ (last column of plots) and other popular visual tracking benchmarks on the distributions computed for different bounding box characteristics. Each column of plots reports the distribution of bounding box sizes, scale changes, and aspect ratio change (the x-axis of each plot reports the range of the bounding box statistic).}
\label{fig:boxdist}
\end{figure*}

\paragraph{Differences With Other Tracking Benchmarks}
We believe that the proposed \datasetname\ benchmark dataset offers \emph{complementary} features with respect to the existing visual tracking benchmarks.

Table \ref{tab:datasets} and Figure \ref{fig:attrdistributions}(a-b) of the main paper show that \datasetname\ provides complementary characteristics to what is available today to study the performance of visual trackers. Particularly, our proposed dataset offers different distributions of the common challenging factors encountered in other datasets. For example, \datasetname\ includes a larger number of examples with occlusions (POC), fast motion (FM), scale change (SC), aspect ratio change (ARC), illumination variation (IV), and motion blur (MB), while it provides a competitive number of scenarios for low resolution (LR), full occlusion (FOC), deformable objects (DEF), and presence of similar objects (SOB). Additionally, even though the 4 new attributes high resolution (HR), head motion (HM), one-hand interaction (1H),  two-hands interaction (2H), define particular FPV scenarios, we think that they can be of interest even for the visual tracking community. For example, as shown by the second row of images of Figure \ref{fig:annoquality}, 1H and 2H can be considered as attributes that define different levels of occlusion, as objects manipulated with two hands generally cause more extended hiding of the targets. 
Besides these sequence-level features, \datasetname\ offers up to 34 target categories which, to the best of our knowledge, have never been studied. As shown by the Figures \ref{fig:annoquality} and \ref{fig:hardex}, these objects have challenging appearances (e.g. transparent or reflective objects like lids, bottles, or food boxes) and shapes (e.g. knives, spoons, cut food) that change dramatically due to the interaction or motion induced by the camera wearer.

We additionally computed some statistics on the bounding box annotations for the targets in \datasetname\ and on those of other popular visual object tracking benchmarks to understand whether FPV offers different motions of the objects. %
As highlighted by Figure \ref{fig:boxdist}, our dataset exhibits different distributions and thus offers different behaviors of the target appearances and motions. 
Observing the top plot of the last column, it can be noted that \datasetname\ has a wider distribution of bounding box dimensions, making it suitable for the evaluation of trackers with targets of many different sizes. Particularly, \datasetname\ has a larger number of bounding boxes with greater dimension. The plot just below shows that \datasetname\ provides more annotations to assess the capabilities in the tracking of objects that become smaller. Finally, the last plot shows a wider distribution for the aspect ratio change, telling that \datasetname\ offers a large variety of examples to evaluate the capabilities of trackers in predicting the shape change of objects. 

Additionally to these characteristics, we think \datasetname\ is interesting because it allows the study of visual object tracking in unconstrained scenarios of \emph{every-day} situations.

\section{Trackers}

\subsection{\rev{Details of the TbyD-F/H Baselines}}
\label{sec:tbydsupp}
\rev{In this section, we provide the details of the other TbyD-F/H  versions whose results are presented in Table \ref{tab:tbyd}. For the baseline that uses SORT \cite{SORT} (whose results are given in rows 1 and 4 of Table \ref{tab:tbyd}), we applied such bounding box association method over the detections of the underlying detector. Specifically, at the first frame we initialize a Kalman filter for the initial target object's bounding box and assign the ID predicted by this filter as 0. Then, at every subsequent frame, we first obtain all the object detections by inputting the detector with the frame. Such localizations are paired with the respective predicted confidence scores and are given as input to SORT. The latter associates the new detections with the memorized tracklets and, through Kalman filters, refines the localizations for all the objects observed in the scene. Since our problem requires the localization of a single target object (SOT), we return the bounding box associated to ID 0 as the output of this TbyD-F/H version.}

\rev{The baseline whose results are given in rows 3 and 6 of Table \ref{tab:tbyd} uses a combination of the strategy based on the previously predicted bounding box and SORT \cite{SORT}. In more detail, at the first frame of a sequence, we initialize a Kalman filter with the bounding box for the target of interest and assign the ID as 0. The given bounding box is also memorized as the previous target position. At every other frame, we first run the detector and among all the given detections, we retain the one having the largest IoU with the memorized bounding box. This bounding box is paired with the respective score given by the detector, and the resulting concatenated vector is inputted to SORT which then provides the refined bounding box for the target. If the detector does not provide detections or no box has an IoU greater than zero with the memorized box, the latter is used as output localization and SORT is inputted with an empty set of detections/scores.}

\section{\rev{Evaluation}}
\rev{In this section, we describe the additional evaluation protocols and metrics used in our study.}

\paragraph{\rev{Real-Time Evaluation}} 
\label{sec:rte}
\rev{Since many FPV tasks such as object interaction \cite{damen2016you} and early action recognition \cite{RULSTMiccv}, or action anticipation \cite{EK55}, require real-time computation, we evaluate trackers in such a setting by following the details given in \cite{VOT2017,Li2020}.}
\rev{This protocol, which we refer to as RTE, is similar to OPE. A tracker is initialized with the ground-truth in the first frame of a sequence. Then the algorithm is presented with a new frame only after its execution over the previous frame has finished. The new presented frame is the last frame available for the time instant in which the tracker becomes ready to be executed, considering that frames occur regularly based on the frame rate of the video. In other words, all the frames occurring in the time interval between the start and end time instants of the tracker's execution are skipped. For all such frames, the last box given by the tracker is used as location for the target. The overall performance scores (SS, NPS, GSR) are ultimately obtained as for the OPE protocol.}
\rev{Together with those values, we evaluate the trackers' processing speed in frames per second (FPS) to quantify their efficiency.}

\section{\rev{Results}}
\label{sec:addres}

\rev{In this section, we provide complete results for all the set of 42 trackers analyzed in this study. We also report on the real-time performance of trackers and give additional insights on some of the deep learning-based methodologies considered in our work.}

\subsection{\rev{Complete Results}}
\begin{figure*}[!h]%
\centering
\includegraphics[width=1.4\columnwidth]{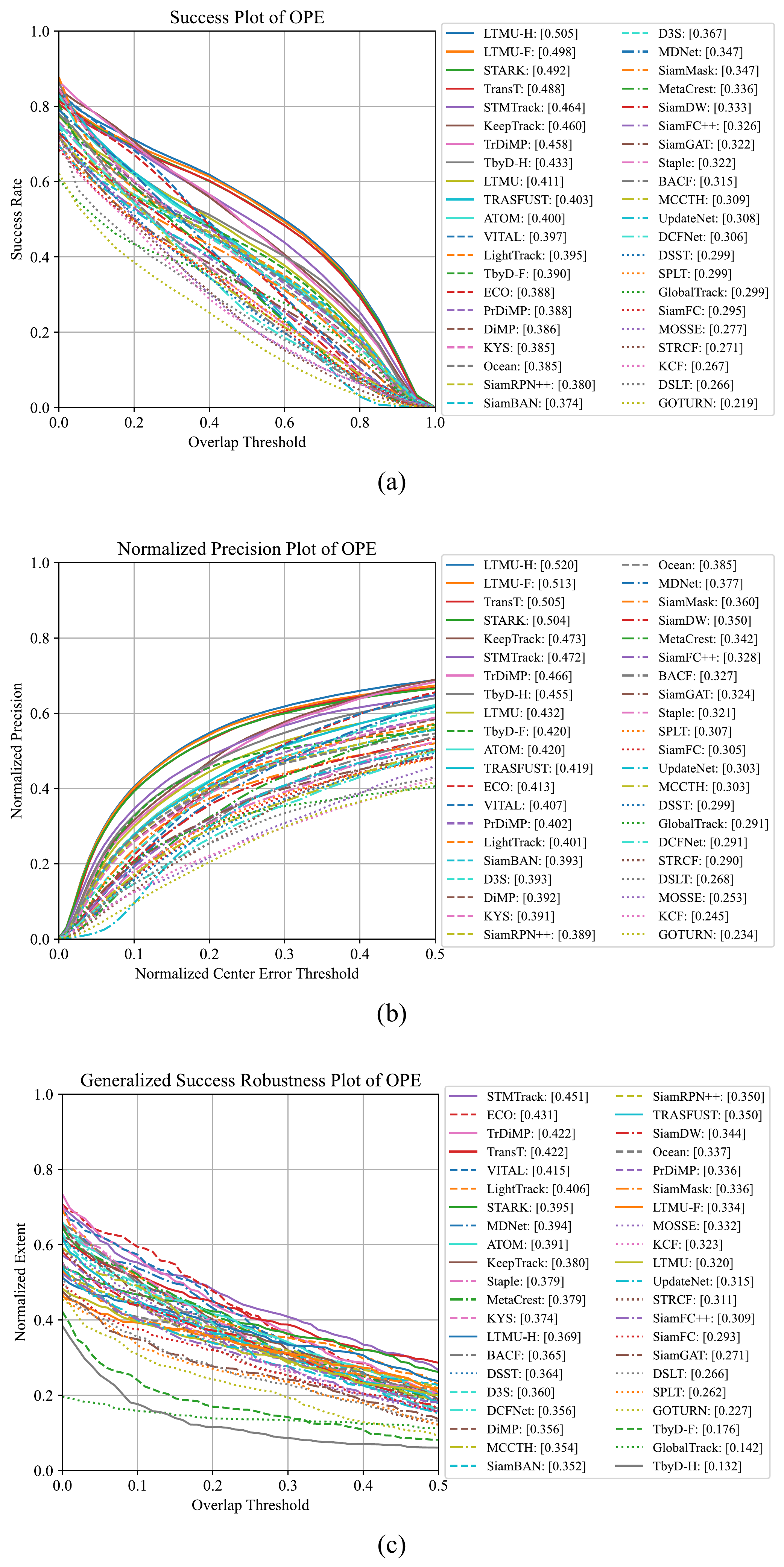}
\caption{Performance of all the 42 selected trackers on the proposed \datasetname\ benchmark under the OPE protocol. \AF{In brackets, next to the trackers' names, we report the SS, NPS, and GSR values.}}
\label{fig:results42}
\end{figure*}

\begin{figure*}[h]%
\centering
\includegraphics[angle=270,width=.35\linewidth]{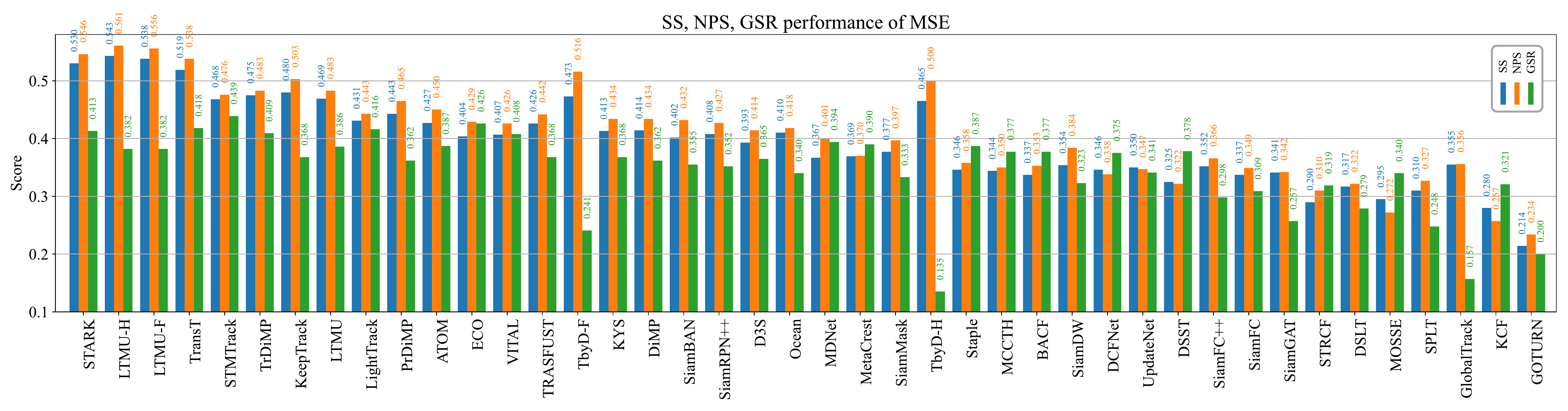}
\caption{SS, NPS, and GSR performance of the 42 benchmarked generic object trackers on the proposed \datasetname\ benchmark achieved under the MSE protocol. The trackers are ordered by the average value of their SS, NPS, GSR scores.}
\label{fig:resultsmse42}
\end{figure*}

\begin{figure*}[h]%
\centering
\includegraphics[width=\linewidth]{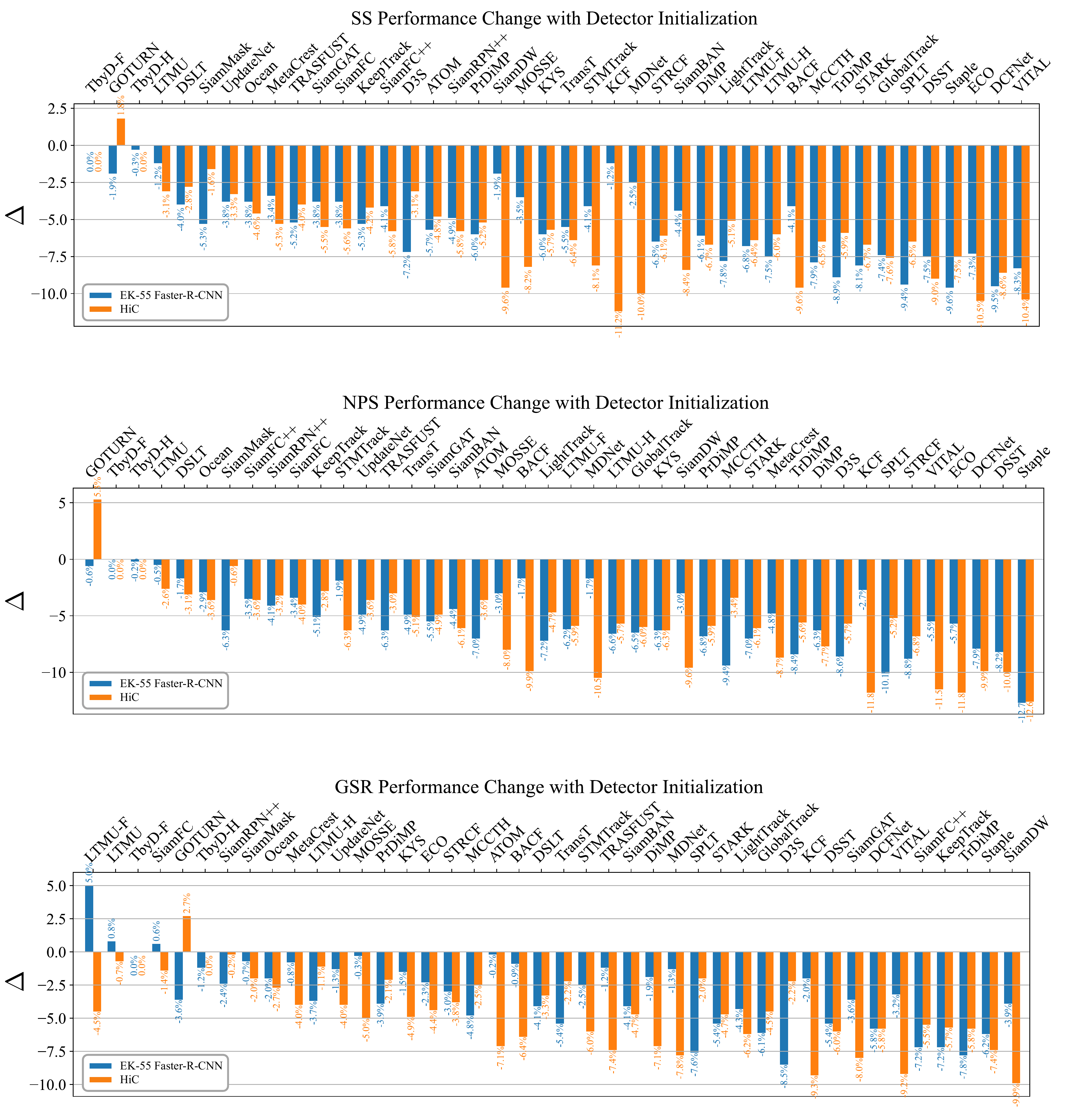}
\caption{\rev{Results of the OPE-D experiment in which the bounding box for initialization is given either by the EK-55 trained Faster-R-CNN \cite{EK55} or the HiC detector \cite{Shan2020}. The performance change with respect with the ground-truth initialization is reported for the SS, NPS, and GSR metrics. In each plot, the trackers are ordered by the average performance change of the two initialization experiments. }}
\label{fig:oped42}
\end{figure*}

\begin{figure*}[!ht]%
\centering
\includegraphics[width=.7\linewidth]{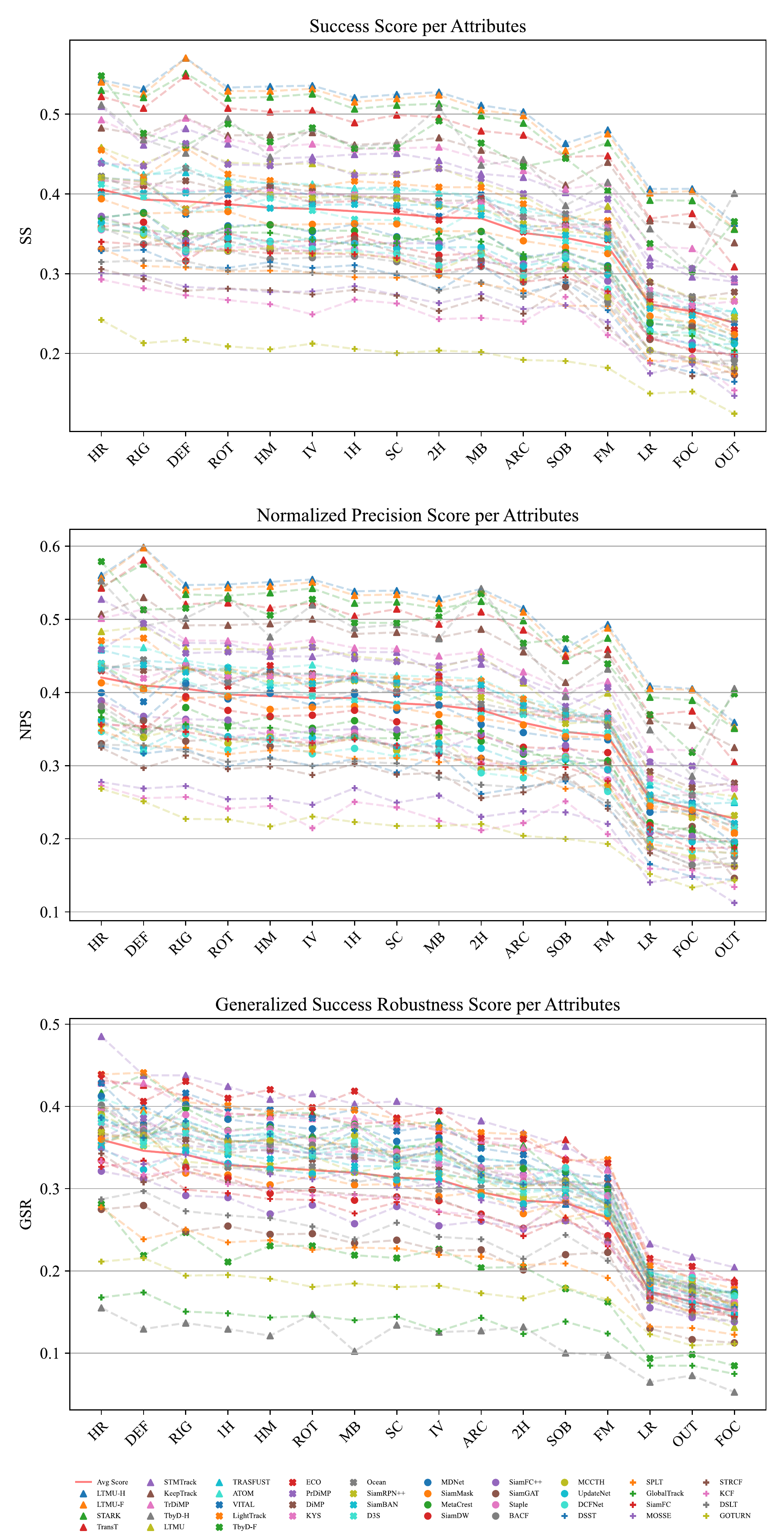}
\caption{SS, NPS, and GSR performance achieved under the MSE protocol of the 42 selected trackers with respect to the sequence attributes available in \datasetname. (The results for the POC attribute are not reported because this attribute is present in every sequence). The red plain line highlights the average tracker performance.}
\label{fig:resattributes42}
\end{figure*}

\begin{figure*}[!ht]%
\centering
\includegraphics[width=.7\linewidth]{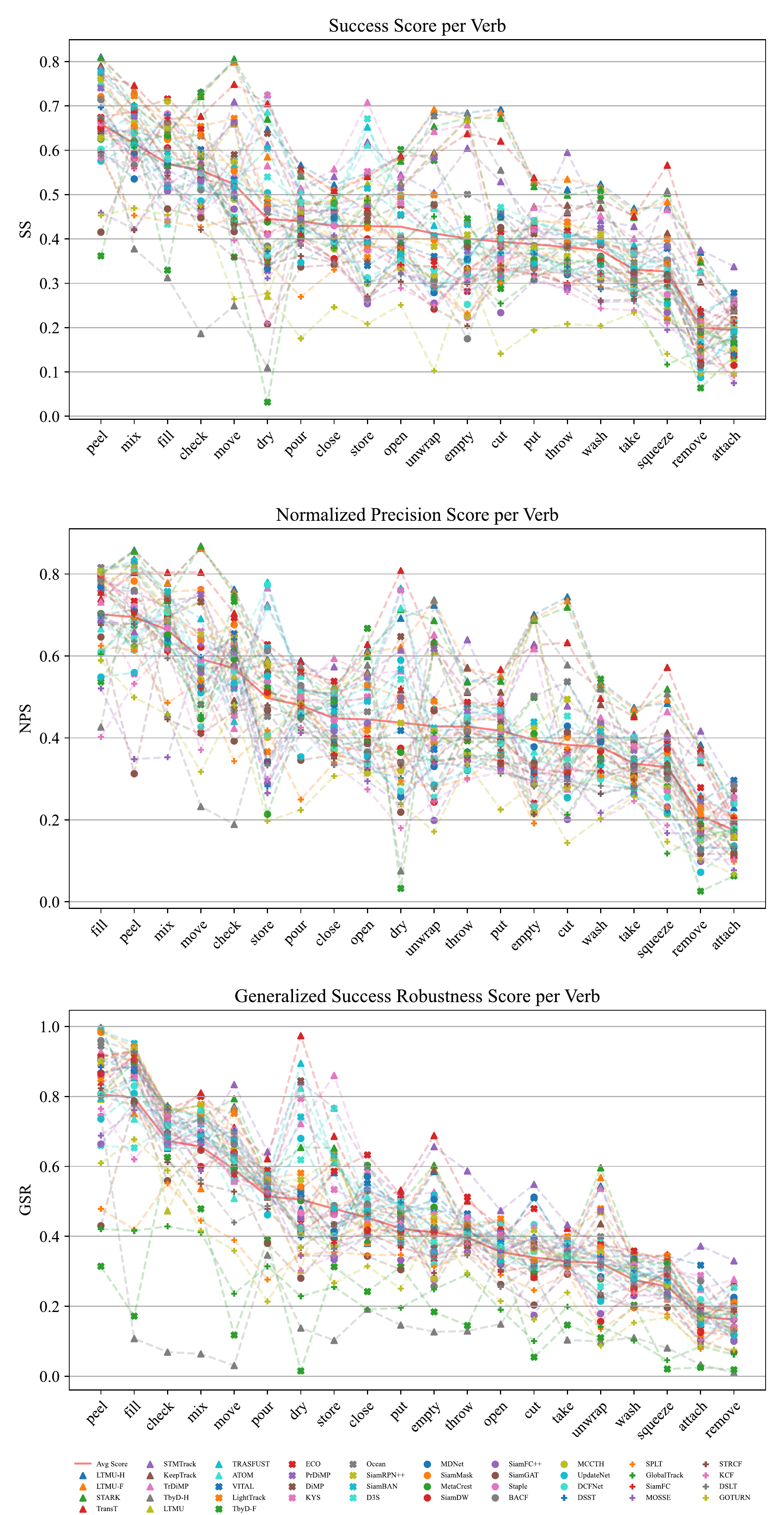}
\caption{SS, NPS, and GSR performance achieved under the MSE protocol of the 42 selected trackers with respect to the sequence verbs available in \datasetname. The red plain line highlights the average tracker performance.}
\label{fig:resverbs42}
\end{figure*}

\begin{figure*}[!ht]%
\centering
\includegraphics[width=.7\linewidth]{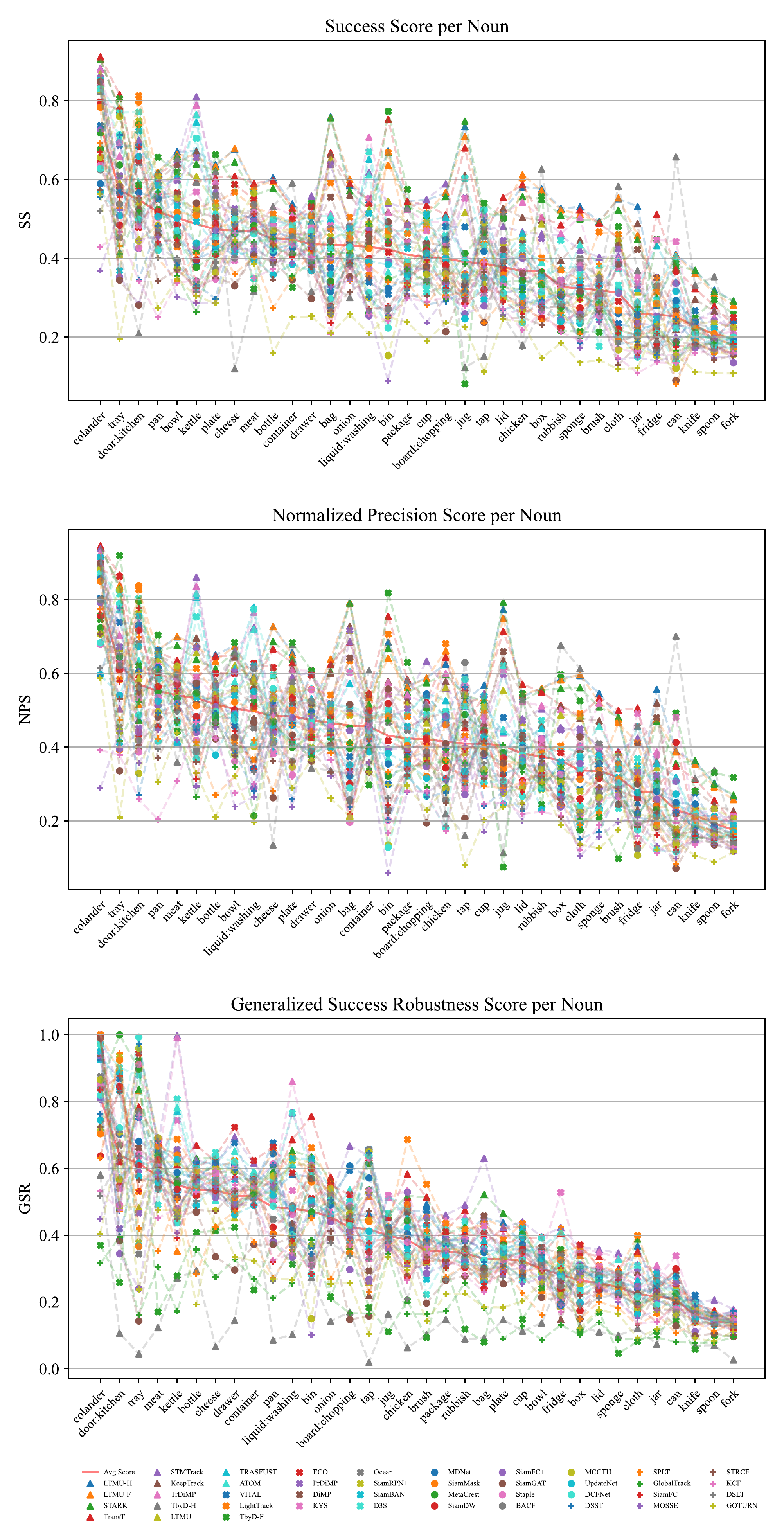}
\caption{SS, NPS, and GSR performance achieved under the MSE protocol of the 42 selected trackers with respect to the target nouns available in \datasetname. The red plain line highlights the average tracker performance.}
\label{fig:resnouns42}
\end{figure*}

\rev{In Figure \ref{fig:results42}, the OPE-based SS, NPS, GSR plots and scores are given for all the 42 considered trackers. Figure \ref{fig:resultsmse42} reports the MSE-based scores, while Figures \ref{fig:resattributes42}, \ref{fig:resverbs42}, \ref{fig:resnouns42} show the MSE-based scores with respect to attributes, verbs, and nouns respectively. Figure \ref{fig:oped42} includes the change in SS, NPS, GSR after the trackers' initialization with object detectors \cite{EK55,Shan2020}, while Tables \ref{tab:handsobjhicfull} and \ref{tab:handsobjgtfull} report the results of the HOI evaluation with the HiC-based pipeline (Table \ref{tab:handsobjhicfull}) and with the oracle-based pipeline (Table \ref{tab:handsobjgtfull}). }

\begin{figure*}[h]%
\centering
\includegraphics[width=\linewidth]{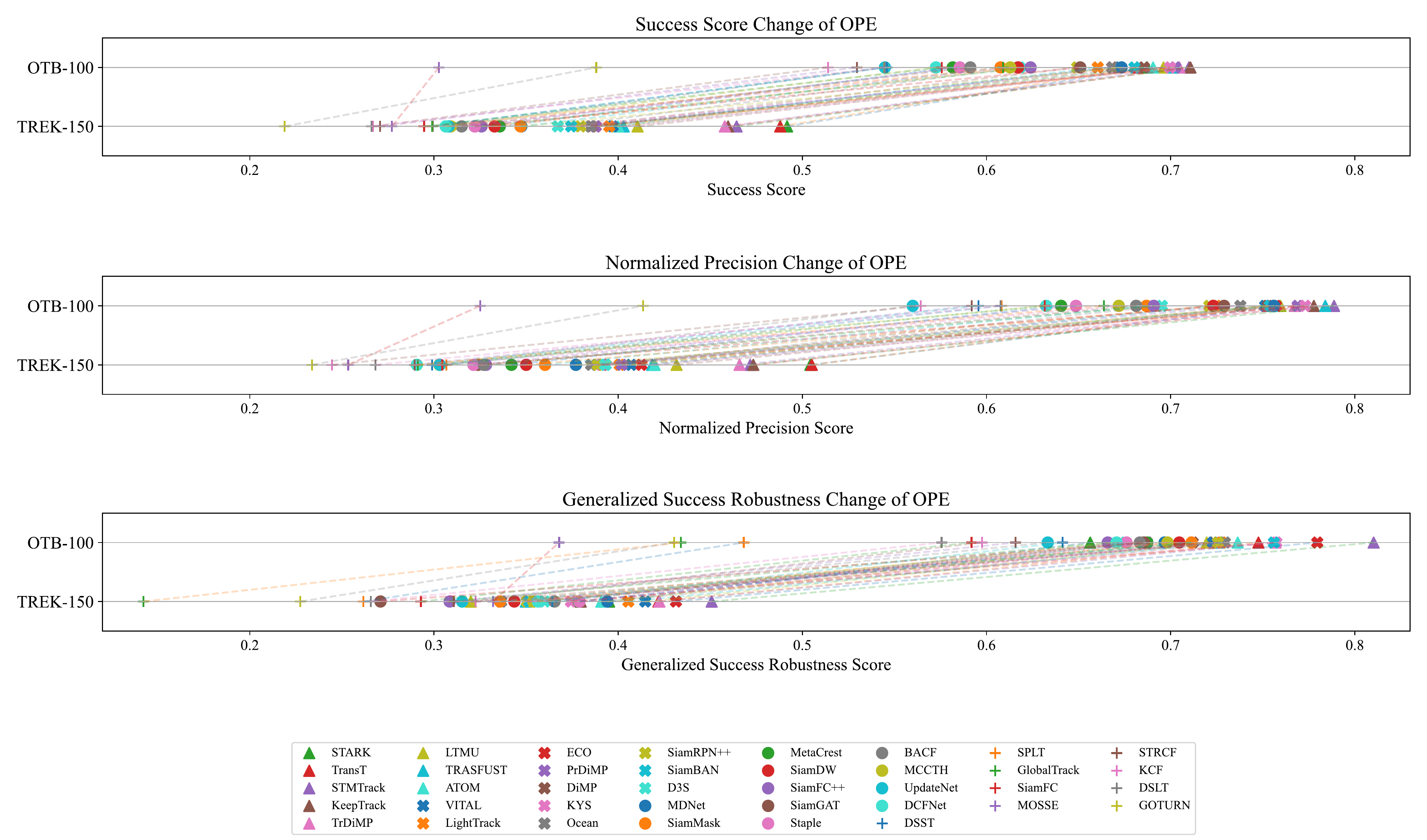}
\caption{Performance comparison of SS, NPS, and GSR scores obtained by the 38 benchmarked generic object trackers on the popular OTB-100 benchmark \cite{OTB} and on the proposed \datasetname\ benchmark under the OPE protocol.}
\label{fig:opechange}
\end{figure*}

\begin{figure*}[t]%
\centering
\includegraphics[width=\linewidth]{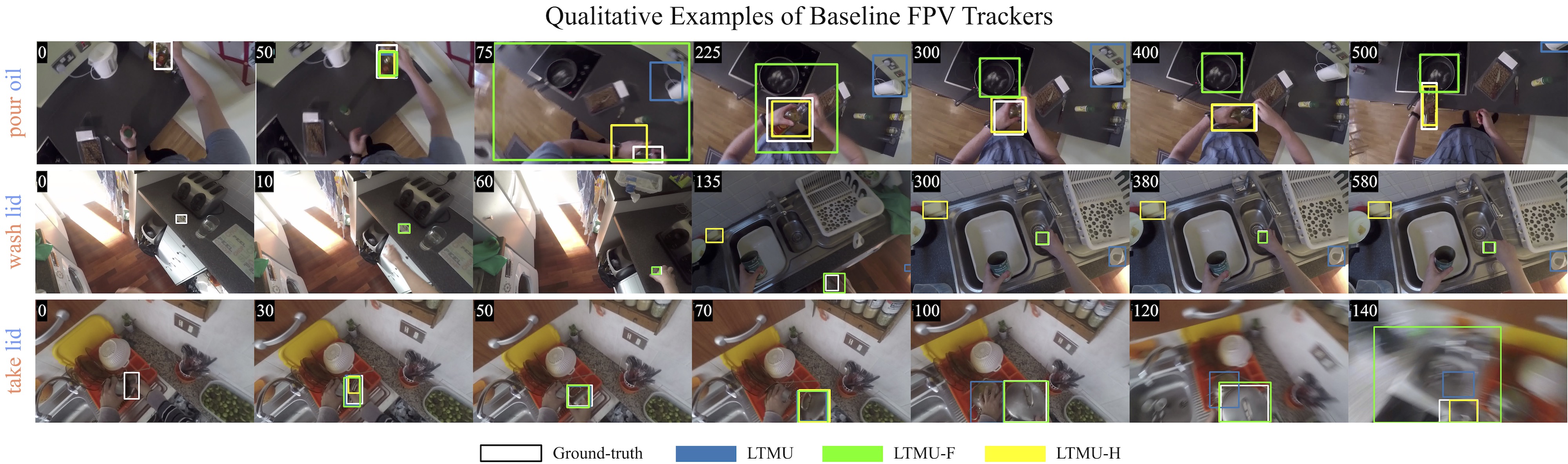}
\caption{Qualitative results of the baseline FPV trackers LTMU-F and LTMU-H in comparison with LTMU.}
\label{fig:qualitativefpv}
\end{figure*}

\begin{table}[t]
\fontsize{8}{9}\selectfont
	\centering
	\caption{Results of the experiment in which trackers are evaluated by the Recall of an FPV HOI detection pipeline where trackers are used as localization method for the object involved in the interaction. The first column presents the results of the proposed system in which each tracker is initialized with the bounding box given by HiC in its first valid HOI detection. The last column reports the SS, NPS, and GSR results achieved by each tracker with the OPE protocol on the sub-sequences yielded by the HOI labels. Best results, per measure, are highlighted in \tblbest{gold}, second-best in \tblsecondbest{silver}, third-best in \tblthirdbest{bronze}.}
	\label{tab:handsobjhicfull}
	\setlength\tabcolsep{.5cm}
	\rowcolors{1}{tblrowcolor1}{tblrowcolor2}
	\begin{tabular}{l | c c }
		\toprule
		Tracker & Recall & (SS, NPS, GSR) \\
		
		\midrule
		
		STARK & \tblbest{0.248} & (\tblbest{0.211}, \tblthirdbest{0.221}, \tblthirdbest{0.222}) \\
		LTMU-H & \tblsecondbest{0.246} & (\tblsecondbest{0.210}, \tblsecondbest{0.222}, 0.217) \\ 
		LTMU-F & \tblthirdbest{0.245} & (\tblsecondbest{0.210}, \tblthirdbest{0.221}, 0.216) \\
		TransT & 0.243 & (\tblthirdbest{0.208}, \tblthirdbest{0.221}, \tblbest{0.229})  \\
		LTMU & 0.241 & (0.204, 0.220, 0.208) \\ 
		TrDiMP & 0.239 & (0.205, 0.213, \tblthirdbest{0.222}) \\ 
		TbyD-H & 0.238 & (0.205, \tblbest{0.223}, 0.163) \\
		LightTrack & 0.233 & (0.197, 0.212, \tblsecondbest{0.228}) \\
		KeepTrack & 0.232 & (0.201, 0.214, 0.212)  \\
		PrDiMP & 0.228 & (0.194, 0.207, 0.206) \\ 
		SiamRPN++ & 0.227 & (0.191, 0.206, 0.209) \\
		SiamBAN & 0.226 & (0.192, 0.209, 0.209) \\ 

		TbyD-F & 0.220 & (0.184, 0.202, 0.179) \\
		SiamMask & 0.219 & (0.185, 0.198, 0.213) \\
		Ocean & 0.218 & (0.189, 0.203, 0.207) \\ 
		STMTrack & 0.216 & (0.196, 0.202, 0.219)   \\
		KYS & 0.212 & (0.187, 0.199, 0.218) \\
		D3S & 0.211 & (0.187, 0.199, 0.208) \\  
		ECO & 0.211 & (0.181, 0.196, 0.217) \\ 
		DiMP & 0.210 & (0.186, 0.198, 0.211) \\
		SiamDW & 0.209 & (0.179, 0.198, 0.208) \\
		MetaCrest & 0.209 & (0.178, 0.188, 0.212) \\ 
		TRASFUST & 0.208 & (0.188, 0.199, 0.210) \\ 
		ATOM & 0.207 & (0.186, 0.198, 0.213) \\ 
		VITAL & 0.198 & (0.178, 0.192, 0.213) \\ 
		SiamFC++ & 0.198 & (0.173, 0.185, 0.184) \\
		SiamFC & 0.195 & (0.171, 0.180, 0.195) \\ 
		GlobalTrack & 0.195 & (0.170, 0.180, 0.144) \\ 
		BACF & 0.188 & (0.170, 0.189, 0.206) \\
		MDNet & 0.185 & (0.170, 0.185, 0.207) \\ 
		DSLT & 0.185 & (0.165, 0.179, 0.188) \\ 
		Staple & 0.182 & (0.164, 0.179, 0.204) \\  
		SiamGAT & 0.182 & (0.162, 0.167, 0.181) \\ 
		UpdateNet & 0.182 & (0.158, 0.158, 0.185) \\ 
		MCCTH & 0.179 & (0.165, 0.172, 0.199) \\ 
		DCFNet & 0.179 & (0.164, 0.171, 0.200) \\ 
		DSST & 0.179 & (0.163, 0.167, 0.203) \\  
		SPLT & 0.178 & (0.155, 0.168, 0.166) \\
		STRCF & 0.169 & (0.161, 0.178, 0.193) \\ 
		KCF & 0.166 & (0.154, 0.159, 0.190) \\ 
		MOSSE & 0.158 & (0.151, 0.154, 0.188) \\ 
		GOTURN & 0.139 & (0.138, 0.147, 0.162) \\

		\bottomrule	
		
\end{tabular}
\end{table}

\begin{table}[t]
\fontsize{8}{9}\selectfont
	\centering
	\caption{Results of the experiment in which trackers are evaluated by the Recall of an FPV HOI detection pipeline where trackers are used as localization method for the object involved in the interaction. The first column presents the results of the proposed system in which each tracker is initialized with the ground-truth bounding box for the first frame labeled with the HOI. The last column reports the SS, NPS, and GSR results achieved by each tracker with the OPE protocol on the sub-sequences yielded by the HOI labels. Best results, per measure, are highlighted in \tblbest{gold}, second-best in \tblsecondbest{silver}, third-best in \tblthirdbest{bronze}.}
	\label{tab:handsobjgtfull}
	\setlength\tabcolsep{.5cm}
	\rowcolors{1}{tblrowcolor1}{tblrowcolor2}
	\begin{tabular}{l | c c }
		\toprule
		Tracker & Recall & (SS, NPS, GSR) \\
		
		\midrule
		LTMU-H & \tblbest{0.754} & (\tblbest{0.648}, \tblbest{0.680}, 0.666) \\
		STARK & \tblsecondbest{0.750} & (\tblsecondbest{0.646}, \tblsecondbest{0.677}, 0.695) \\
		LTMU-F & \tblthirdbest{0.746} & (\tblthirdbest{0.641}, \tblthirdbest{0.672}, 0.667) \\
		TransT & 0.725 & (0.631, 0.659, \tblthirdbest{0.708})  \\
	    STMTrack & 0.671 & (0.595, 0.610, \tblbest{0.726})   \\ 
		LightTrack & 0.670 & (0.586, 0.611, \tblsecondbest{0.710}) \\
		TrDiMP & 0.669 & (0.590, 0.606, 0.688) \\
		LTMU & 0.663 & (0.578, 0.613, 0.623) \\ 
		KeepTrack & 0.661 & (0.587, 0.611, 0.672)  \\

		SiamRPN++ & 0.660 & (0.577, 0.605, 0.678) \\ 
		PrDiMP & 0.653 & (0.573, 0.600, 0.669) \\ 
		SiamBAN & 0.637 & (0.566, 0.603, 0.666) \\ 
		TRASFUST & 0.617 & (0.562, 0.594, 0.659) \\  
		TbyD-F & 0.617 & (0.528, 0.578, 0.509) \\
		SiamMask & 0.615 & (0.542, 0.576, 0.643) \\ 
		ATOM & 0.614 & (0.559, 0.590, 0.670) \\ 
		ECO & 0.613 & (0.556, 0.600, 0.701) \\ 
		VITAL & 0.610 & (0.553, 0.580, 0.682) \\
		KYS & 0.609 & (0.559, 0.575, 0.660) \\
		Ocean & 0.608 & (0.538, 0.560, 0.649) \\
		DiMP & 0.607 & (0.547, 0.574, 0.650) \\
		TbyD-H & 0.603 & (0.544, 0.582, 0.389) \\
		D3S & 0.598 & (0.541, 0.566, 0.638) \\ 
		SiamDW & 0.572 & (0.512, 0.552, 0.642) \\
		MetaCrest & 0.569 & (0.518, 0.535, 0.662) \\
		MDNet & 0.570 & (0.518, 0.560, 0.664) \\ 
		GlobalTrack & 0.563 & (0.493, 0.507, 0.440) \\ 
		
		SiamFC & 0.549 & (0.504, 0.530, 0.617) \\ 
		MCCTH & 0.547 & (0.510, 0.529, 0.650) \\ 
		Staple & 0.544 & (0.510, 0.537, 0.655) \\  
		SiamFC++ & 0.542 & (0.495, 0.519, 0.602) \\
		UpdateNet & 0.540 & (0.470, 0.473, 0.612) \\
		BACF & 0.536 & (0.507, 0.551, 0.665) \\
		DCFNet & 0.531 & (0.508, 0.523, 0.651) \\ 
		DSST & 0.531 & (0.503, 0.518, 0.653) \\ 
		DSLT & 0.514 & (0.481, 0.507, 0.587) \\
		KCF & 0.488 & (0.472, 0.482, 0.622) \\ 
		SiamGAT & 0.487 & (0.465, 0.468, 0.533) \\ 
		STRCF & 0.481 & (0.466, 0.501, 0.612) \\ 
		SPLT & 0.480 & (0.450, 0.473, 0.541) \\
		MOSSE & 0.459 & (0.459, 0.459, 0.614) \\ 
		GOTURN & 0.367 & (0.384, 0.404, 0.509) \\

		\bottomrule	
		
\end{tabular}
\end{table}

\begin{table}[!ht]
\fontsize{8}{9}\selectfont
	\centering
	\caption{Performance achieved by the 42 trackers benchmarked on \datasetname\ using the RTE protocol. Best results, per measure, are highlighted in \tblbest{gold}, second-best in \tblsecondbest{silver}, third-best in \tblthirdbest{bronze}.}
	\label{tab:realtime32}
	\setlength\tabcolsep{.4cm}
	\tblalternaterowcolors
	\begin{tabular}{l | c c c c }
		\toprule
		
		Tracker & FPS & SS & NPS & GSR  \\

		\midrule
		TransT & 19 & \tblbest{0.462} & \tblbest{0.471} & \tblsecondbest{0.394} \\
		STARK & 14 & \tblsecondbest{0.453} & \tblsecondbest{0.456} & \tblthirdbest{0.345} \\
		STMTrack & 13 & \tblthirdbest{0.434} & \tblthirdbest{0.440} & \tblbest{0.407} \\
		TrDiMP & 9 & 0.389 & 0.378 & 0.287 \\
		LightTrack & 8 & 0.376 & 0.373 & 0.335 \\
		Ocean & 21 & 0.365 & 0.358 & 0.294 \\
		SiamRPN++ & 23 & 0.362 & 0.356 & 0.293 \\
		SiamBAN & 24 & 0.360 & 0.369 & 0.313 \\
		PrDiMP & 13 & 0.352 & 0.349 & 0.243 \\
		KeepTrack & 9 & 0.345 & 0.335 & 0.188 \\
		DiMP & 16 & 0.336 & 0.331 & 0.224 \\
		SiamMask & 23 & 0.335 & 0.333 & 0.298 \\
		SiamFC++ & \tblsecondbest{45} & 0.330 & 0.331 & 0.308 \\
		SiamDW & 32 & 0.327 & 0.334 & 0.317  \\
		KYS & 12 & 0.327 & 0.317 & 0.219 \\
		ATOM & 15 & 0.319 & 0.312 & 0.179 \\
		SiamGAT & 20 & 0.314 & 0.306 & 0.257 \\
		UpdateNet & 21 & 0.311 & 0.297 & 0.295 \\
		DCFNet & \tblbest{49} & 0.299 & 0.286 & 0.335 \\
		TRASFUST & 13 & 0.296 & 0.270 & 0.185 \\
		SiamFC & 34 & 0.293 & 0.295 & 0.280 \\
		D3S & 16 & 0.276 & 0.263 & 0.182 \\
		BACF & 9 & 0.276 & 0.262 & 0.234 \\
		SPLT & 8 & 0.265 & 0.247 & 0.203 \\
		STRCF & 10 & 0.264 & 0.250 & 0.218 \\
		DSLT & 7 & 0.260 & 0.234 & 0.211 \\
		ECO & 15 & 0.252 & 0.231 & 0.173 \\
		GlobalTrack & 8 & 0.253 & 0.227 & 0.139  \\
		MCCTH & 8 & 0.251 & 0.231 & 0.232 \\
		Staple & 13 & 0.249 & 0.236 & 0.234 \\
		GOTURN & \tblthirdbest{44} & 0.247 & 0.242 & 0.119 \\
		LTMU-H & 9 & 0.243 & 0.205 & 0.163 \\
		MOSSE & 26 & 0.227 & 0.190 & 0.244 \\
		LTMU-F & 7 & 0.222 & 0.180 & 0.162 \\
		LTMU & 3 & 0.213 & 0.178 & 0.161 \\
        MetaCrest & 8 & 0.207 & 0.175 & 0.165 \\
        VITAL & 4 & 0.204 & 0.165 & 0.158 \\
        DSST & 2 & 0.191 & 0.145 & 0.161 \\
        TbyD-F & 1 & 0.191 & 0.135 & 0.163 \\
        KCF & 6 & 0.186 & 0.157 & 0.177 \\
        MDNet & 1 & 0.185 & 0.140 & 0.161 \\
        TbyD-H & 8 & 0.175 & 0.140 & 0.127 \\
		\bottomrule		
\end{tabular}
\end{table}

\subsection{Processing Speed Study}
Table \ref{tab:realtime32} \AF{reports} the FPS performance of the 42 trackers and the SS, NPS, and GSR scores achieved under the RTE protocol.
None of the trackers achieve the frame rate speed of 60 FPS. We argue \AF{that} this is due to the full HD \AF{resolution of} frames \AF{which requires demanding image crop and resize operations with targets of considerable size.}
Taking into consideration the tracking approaches, we observe that trackers based  on single-shot siamese networks (e.g. Ocean, SiamBAN, SiamRPN++) or on light online adaptation techniques such as the target template change (e.g. as performed by TransT, STARK, STMTrack)
emerge as the fastest trackers and exhibit a less significant performance drop of the proposed scores. 
 In particular, the decrease in SS of SiamBAN and SiamRPN++ is of 3.7\% and 4.7\%, respectively, with respect to the OPE results. TransT and STARK exhibit a larger performance drop of 5.3 \% and 6.1\% respectively, but their robustness makes their overall real-time performance much higher than SiamBAN and SiamRPN++.
Due to the reliance on heavier online learning mechanisms, trackers like KeepTrack, PrDiMP, ATOM, KYS, ECO achieve a lower processing
speed that consequently causes a major accuracy loss in real-time scenarios. 
\rev{The proposed FPV tracking baselines show a consistent drop in real-time settings. This is due to their reliance on demanding models such as the FPV object detectors \cite{EK55,Shan2020}. LTMU-F/H result a bit better on this point since they do not execute the detector at every frame, but only when triggered by the target verification mechanism. The performance drop is particularly large with respect the underlying tracker STARK because of the long initialization time (that lasts 8 seconds on average) that LTMU-F/H takes to create all the models involved in the pipeline. 
TbyD-F/H execute the time-consuming detectors at every frame, resulting in an even lower performance due to the many frames skipped during the tracking.}

In general, we observe that the GSR score is the measure on which all trackers present the major performance drop in the real-time setting, suggesting that particular effort should be spent to make trackers better address longer references to objects in real-time scenarios.
Overall, we can say that trackers like TransT and STARK are currently the most suitable methods to employ for the development of real-time FPV applications requiring object tracking. Given their limited performance decrease between the OPE and RTE results, siamese-based trackers could serve as promising alternatives if their tracking accuracy and robustness are improved.

\subsection{\rev{Study of Deep Learning Trackers}}
\label{sec:fpvtrain}
\rev{Considering that nowadays most of the state-of-the-art solutions are based on deep neural networks, in this section we provide insights about how the FPV performance of such methods depends on the training data and on the neural network design. In these experiments, we consider the trackers SiamRPN++, DiMP, and STARK, as the representative methods of the popular and successful approaches of siamese networks, deep discriminative correlation filters, and transformers.}

\begin{table}[t]
\fontsize{8}{9}\selectfont
	\centering
	\caption{\rev{OPE-based performance of the deep learning-based trackers SiamRPN++, DiMP, STARK on \datasetname\ after being trained separately on each of the most popular generic object TPV datasets (TrackingNet, GOT-10k, LaSOT). 
	} }
	\label{tab:trainset}
	\begin{tabular}{c | c | c  | c | c }
		\toprule
		\multirow{2}{*}{Tracker} & \multirow{2}{*}{Metric} & \multicolumn{3}{c}{Training Dataset} \\
                   & & TrackingNet & GOT-10k & LaSOT \\
		\midrule
		
		\multirow{3}{*}{SiamRPN++} & SS & 0.307 & 0.332 & 0.281 \\
		& NPS & 0.315 & 0.343 & 0.290 \\
		& GSR & 0.298 & 0.318 & 0.280 \\
		
		\midrule
		
		\multirow{3}{*}{DiMP} & SS & 0.349 & 0.322 & 0.378 \\
		& NPS & 0.349 & 0.327 & 0.396 \\
		& GSR & 0.331 & 0.291 & 0.343 \\
		
		\midrule
		
		\multirow{3}{*}{STARK} & SS & 0.314 & 0.315 & 0.340 \\
		& NPS & 0.314 & 0.304 & 0.332 \\
		& GSR & 0.244 & 0.182 & 0.199 \\

		\bottomrule		
\end{tabular}
\end{table}

\paragraph{\rev{Impact of Training Data}}
\rev{As reported in Table \ref{tab:trackers}, deep learning-based trackers use several different datasets to optimize the neural network-based modules used inside their processing pipelines. At a first glance, it is not easy to correlate the tracking performance in FPV with the training set reported in such a table. To overcome this problem, we re-trained the 3 representative trackers SiamRPN++, DiMP, STARK, on the training set of each of the most popular large-scale tracking datasets currently available, i.e. TrackingNet, GOT-10k, and LaSOT. These datasets comprise videos of generic objects acquired from a third person view (TPV) generally. Despite they provide similar characteristics in the video perspective and target's nature, such datasets have different features, especially in the size and the object categories. For example, TrackingNet offers a training set of more than 30K videos and 14M frames with 21 target object categories, GOT-10k provides 9.3K videos for a total of 1.4M frames and 480 object categories, while LaSOT's training set comprises 1.1K videos, 2.8M frames, and 70 object categories. }
\rev{To train the trackers, we followed the instructions given in the original papers and code repositories (which were publicly available). We kept the same hyperparameters for all the trackers and swapped only the training dataset. For the STARK tracker, we modified only the number of training epochs to a total of 20 (for both the two training stages) in order to achieve a reasonable training time on our hardware. We found such a number of epochs to be sufficient for achieving a significant tracking performance allowing the emergence of the performance difference after training on the different data distributions.}
\rev{The performance achieved by such trackers on TREK-150 after being trained with the aforementioned training sets is reported in Table \ref{tab:trainset}. The relevant thing to understand here is how, for every methodology,  the performance changes as the training set is changed. Siamese network-based trackers (SiamRPN++-like) benefit the most from the GOT-10k training set, suggesting that the large amount of object categories available in this dataset helps in making the tracking better generalize to the objects and scenarios present in FPV. 
Deep discriminative (DiMP-like) and transformer trackers (STARK-like) instead benefit the most from the LaSOT training set, suggesting that these methodologies need a balance between the number of object categories and the amount of appearance change in time (LaSOT provides long videos in which targets are subject to severe appearance changes) to perform well in FPV.
Overall, we can observe the highest FPV tracking performance is not achieved with the largest number of videos and frames as available in TrackingNet. These results demonstrate that a large number of videos/frames is not sufficient to perform the best in FPV, but a more carefully designed training set should be used instead. }

\begin{table}[t]
\fontsize{8}{9}\selectfont
	\centering
	\caption{\rev{Performance of the deep learning-based trackers SiamRPN++, DiMP, STARK on \datasetname\ when executed with different CNN backbones. The trackers have been executed under the OPE protocol. 
	} }
	\label{tab:backbone}
	\setlength\tabcolsep{.1cm}
	\begin{tabular}{c | c | c  | c | c | c }
		\toprule
		\multirow{2}{*}{Tracker} & \multirow{2}{*}{Metric} & \multicolumn{4}{c}{Backbone CNN} \\
                   & & AlexNet & ResNet-18 & ResNet-50 & ResNet-101 \\
		
		\midrule

		\multirow{3}{*}{SiamRPN++} & SS & 0.345 & - & 0.380 & - \\
		& NPS & 0.353 & - & 0.389 & - \\
		& GSR & 0.325 & - & 0.350 & - \\
		
		\midrule

		\multirow{3}{*}{DiMP} & SS & - & 0.373 & 0.386 & -\\
		& NPS & - & 0.383 & 0.392 & - \\
		& GSR & - & 0.339 & 0.356 & - \\

		\midrule

		\multirow{3}{*}{STARK} & SS & - & - & 0.492 & 0.490\\
		& NPS & - & - & 0.504 & 0.505 \\
		& GSR & - & - & 0.395 & 0.389 \\

		\bottomrule		
\end{tabular}
\end{table}

\paragraph{\rev{Impact of the CNN Backbone}}
\rev{Table \ref{tab:backbone} reports the performance of SiamRPN++, DiMP, and STARK trackers configured with different backbone networks. In this experiment, we wanted to understand which is the CNN that produces the best appearance features for FPV-based tracking. We used the pre-trained models provided by the respective authors and trained using the popular TPV tracking benchmarks. As can be easily noticed, for all the methodologies, the ResNet-50 is the backbone CNN that leads to the highest tracking performance in FPV.}

\begin{table}[t]
\fontsize{8}{9}\selectfont
	\centering
	\caption{\rev{Performance of the deep learning trackers SiamRPN++, DiMP, STARK on  \datasetname\ when trained on datasets of generic object tracking (TPV) and on a large-scale dataset for FPV object detection.}}
	\label{tab:ek55od}
	\setlength\tabcolsep{.14cm}
	\begin{tabular}{l | c | c  c  c | c  c  c  }
		\toprule
		\multirow{2}{*}{Tracker} & Training & \multicolumn{3}{c|}{OPE}  & \multicolumn{3}{c}{MSE}\\
                    & Data & SS & NPS & GSR  & SS & NPS  & GSR \\
		\midrule
		\multirow{2}{*}{SiamRPN++} & TPV & 0.380 & 0.389 & 0.350 & 0.408 & 0.427 & 0.352 \\
		& FPV & 0.343 & 0.364 & 0.327 & 0.376 & 0.402 &  0.322 \\
		
		\midrule
		
		\multirow{2}{*}{DiMP} & TPV & 0.386 & 0.392 & 0.357 & 0.414 & 0.434 & 0.362 \\
		& FPV & 0.418 & 0.441 & 0.382 & 0.454 & 0.480 &  0.394 \\
		
		\midrule
		
		\multirow{2}{*}{STARK} & TPV & 0.340 & 0.332 &  0.199 & 0.373 & 0.366 &  0.209 \\
		& FPV & 0.264 & 0.251 & 0.170 & 0.332 & 0.351 &  0.225 \\
		
		\bottomrule		
\end{tabular}
\vspace{3em}
\end{table}
\paragraph{\rev{Deep Learning Trackers Trained for FPV}}
\rev{We also assessed the impact of FPV-specific data to train SiamRPN++, DiMP, and STARK. At the time of writing, the only available data to be exploited for tracking is the subset of EK-55 frames labeled with the bounding boxes of objects for object detection tasks \cite{EK55}. As expressed in Table \ref{tab:trackers}, object detection datasets such as COCO \cite{COCO} are exploited for training the deep models used inside trackers. The idea is to create synthetic videos from still images by applying transformations such as shift, scale, rotation, color jitter, to images of objects. We followed the same approach with EK-55's object detection images and annotations. As training set for the 3 trackers, we considered the subset of images associated to the camera wearers that are not present in TREK-150 (15 people). In total, 197229 frames and 214878 bounding boxes are present in this set for an overall number of 295 object categories. For comparison, COCO's training set has 118287 frames, 860001 bounding boxes belonging to 80 different categories. We retrained each of the trackers on such FPV training set by following the original instructions and hyperparameters. As for the earlier experiment, STARK was trained for 20 epochs in each of the two stages. In Table \ref{tab:ek55od} we present a comparison between the trackers' instances trained for generic object tracking on standard TPV tracking datasets and those trained for FPV (for STARK we consider the best instance presented in Table \ref{tab:trainset} since it was trained with the same hyperparameters). The DiMP tracker is the only one benefiting from FPV-specific data for training. Indeed, its performance is improved by a good margin according to all the metrics. The other trackers, SiamRPN++ and STARK, do not benefit from this data and their tracking performance is weaker than the counterparts trained in TPV. These results suggest that online adaptative methods such as deep discriminative trackers benefit from an initial offline learning stage specifically designed for the domain of application. Less target adaptive trackers such as siamese networks and transformers require large-scale and diverse samples to learn an effective similarity function. Despite the improvement, the FPV-trained DiMP still does not achieve the performance that is observed in domains represented by other benchmarks (e.g. OTB-100, UAV123, LaSOT).  }

\begin{table}[t]
\fontsize{8}{9}\selectfont
	\centering
	\caption{Performance of the offline trackers SiamFC and SiamRPN++ on a subset of 50 sequences of \datasetname\ without and with fine-tuning on the remaining 100 videos.}
	\label{tab:finetuning}
	\setlength\tabcolsep{.14cm}
	\begin{tabular}{l | c | c  c  c | c  c  c  }
		\toprule
		\multirow{2}{*}{Tracker} & Fine- & \multicolumn{3}{c|}{OPE}  & \multicolumn{3}{c}{MSE}\\
                    & tuning & SS & NPS & GSR  & SS & NPS  & GSR \\
		\midrule
		\multirow{2}{*}{SiamFC}&  & 0.311 & 0.332 & 0.317 & 0.307 & 0.317 & 0.307 \\
		& \checkmark & 0.267 & 0.275 & 0.278 & 0.287 & 0.305 & 0.292 \\
		
		\midrule
		
		\multirow{2}{*}{SiamRPN++}&  & 0.384 & 0.395 & 0.377 & 0.367 & 0.385 & 0.333 \\
		& \checkmark & 0.348 & 0.407 & 0.313 & 0.336 & 0.406 & 0.314 \\
		\bottomrule		
\end{tabular}
\vspace{3em}
\end{table}

\paragraph{\rev{Deep Learning Trackers Fine-Tuned for FPV}}
\rev{The previous experiment revealed that large-scale FPV object detection data does not help those trackers} based on siamese neural networks (e.g. SiamFC, SiamRPN++, SiamBAN, SiamGAT). 
\rev{A reason for this could be the fact that even though different transformations are applied to the still images, the appearance of the target does not change so severely as could be in the case of real object manipulation. Hence, we performed additional experiments on such methods }
to understand if their behavior can be improved by learning 
\rev{through object appearances that change through time.}
Our \datasetname\ dataset is designed to evaluate the progress of visual tracking solutions in FPV and does not provide a large-scale database of learning examples as needed by these methods (a large-scale dataset for the training of FPV-specific trackers is out of the scope of this paper). In this view, \datasetname\ well aligns with real-world datasets where millions of frames are not available for training. In such scenarios, the reasonable options the machine learning community suggests are to use the deep learning models as they are because of their general knowledge, or to adapt them through fine-tuning using a smaller training set. 
We carried out the second strategy by randomly splitting \datasetname\ into a training and a test set of 100 and 50 videos respectively. We fine-tuned the popular offline trackers SiamFC and SiamRPN++ on such training set according to their original learning strategy. We then tested the fine-tuned versions on the test set and the results are reported in Table \ref{tab:finetuning} in comparison with the original counterparts. The results show that the simple fine-tuning leads to substantial overfitting that cause the performance to drop in general. The exceptions are given by SiamRPN++'s NPS results which are increased through such adaptation procedure.

\rev{Overall, by considering the outcomes of this experiment and the ones of the previous assessment, as well as the results achieved by trackers trained for generic object tracking, we can conclude that FPV introduces challenging scenarios for tracking methods that cannot be completely addressed by the current availability of tracking data (both TPV and FPV) as well as the knowledge in deep learning-based visual tracking.}

\clearpage
\bibliographystyle{spmpsci}   
\bibliography{snbibcleared}

\end{document}